\documentclass{article}
\usepackage{arxiv}

\usepackage[utf8]{inputenc} 
\usepackage[T1]{fontenc}    
\usepackage{hyperref}       
\usepackage{url}            
\usepackage{booktabs}       
\usepackage{amsfonts}       
\usepackage{nicefrac}       
\usepackage{microtype}      
\usepackage{lipsum}		
\usepackage{graphicx}
\usepackage{doi}
\usepackage{multirow}
\usepackage{multicol}
\usepackage{xcolor}
\usepackage{subcaption}
\usepackage[normalem]{ulem}
\usepackage{amsmath,amsfonts,bm}
\usepackage{amssymb,amsthm}
\usepackage{enumitem}
\usepackage{subcaption}
\usepackage{adjustbox}
\usepackage[
    backend=biber,
    citestyle=authoryear,
    bibstyle=authortitle,
    mincitenames=1,
    maxcitenames=1,
    maxbibnames=3,
]{biblatex}
\newcommand{\citep}[1]{\parencite{#1}}
\setlist[itemize,1]{leftmargin=\dimexpr 18pt}
\setlist[enumerate,1]{leftmargin=\dimexpr 18pt}

\title{
\raisebox{-0.1\height}{\includegraphics[width=0.032\textwidth]{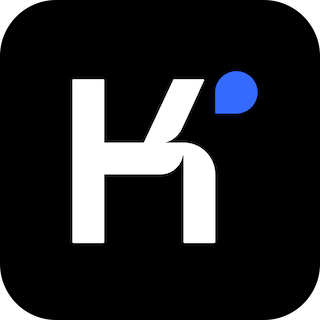}} %
Kimi k1.5: \\Scaling Reinforcement Learning with LLMs}

\author{Kimi Team}

\date{}

\renewcommand{\headeright}{Technical Report}
\renewcommand{\undertitle}{Technical Report of Kimi k1.5}
\renewcommand{\shorttitle}{\raisebox{-0.12\height}{\includegraphics[width=0.02\textwidth]{figures/logo.png}}
Kimi k1.5}

\begin{document}
\maketitle

\vspace{-10pt}
\begin{abstract}
	Language model pretraining with next token prediction has proved effective for scaling compute but is limited to the amount of available training data. Scaling reinforcement learning (RL) unlocks a new axis for the continued improvement of artificial intelligence, with the promise that large language models (LLMs) can scale their training data by learning to explore with rewards. However, prior published work has not produced competitive results. In light of this, we report on the training practice of Kimi k1.5, our latest multi-modal LLM trained with RL, including its RL training techniques, multi-modal data recipes, and infrastructure optimization. 
    Long context scaling and improved policy optimization methods are key ingredients of our approach, which establishes a simplistic, effective RL framework without relying on more complex techniques such as Monte Carlo tree search, value functions, and process reward models.
    Notably, our system achieves state-of-the-art reasoning performance across multiple benchmarks and modalities---e.g., 77.5 on AIME, 96.2 on MATH 500, 94-th percentile on Codeforces, 74.9 on MathVista---matching OpenAI's o1. Moreover, we present effective long2short methods that use long-CoT techniques to improve short-CoT models, yielding state-of-the-art short-CoT reasoning results---e.g., 60.8 on AIME, 94.6 on MATH500, 47.3 on LiveCodeBench---outperforming existing short-CoT models such as GPT-4o and Claude Sonnet 3.5 by a large margin (up to +550\%).
\end{abstract}


\begin{figure}[htb]
    \centering
    \includegraphics[width=0.9\textwidth]{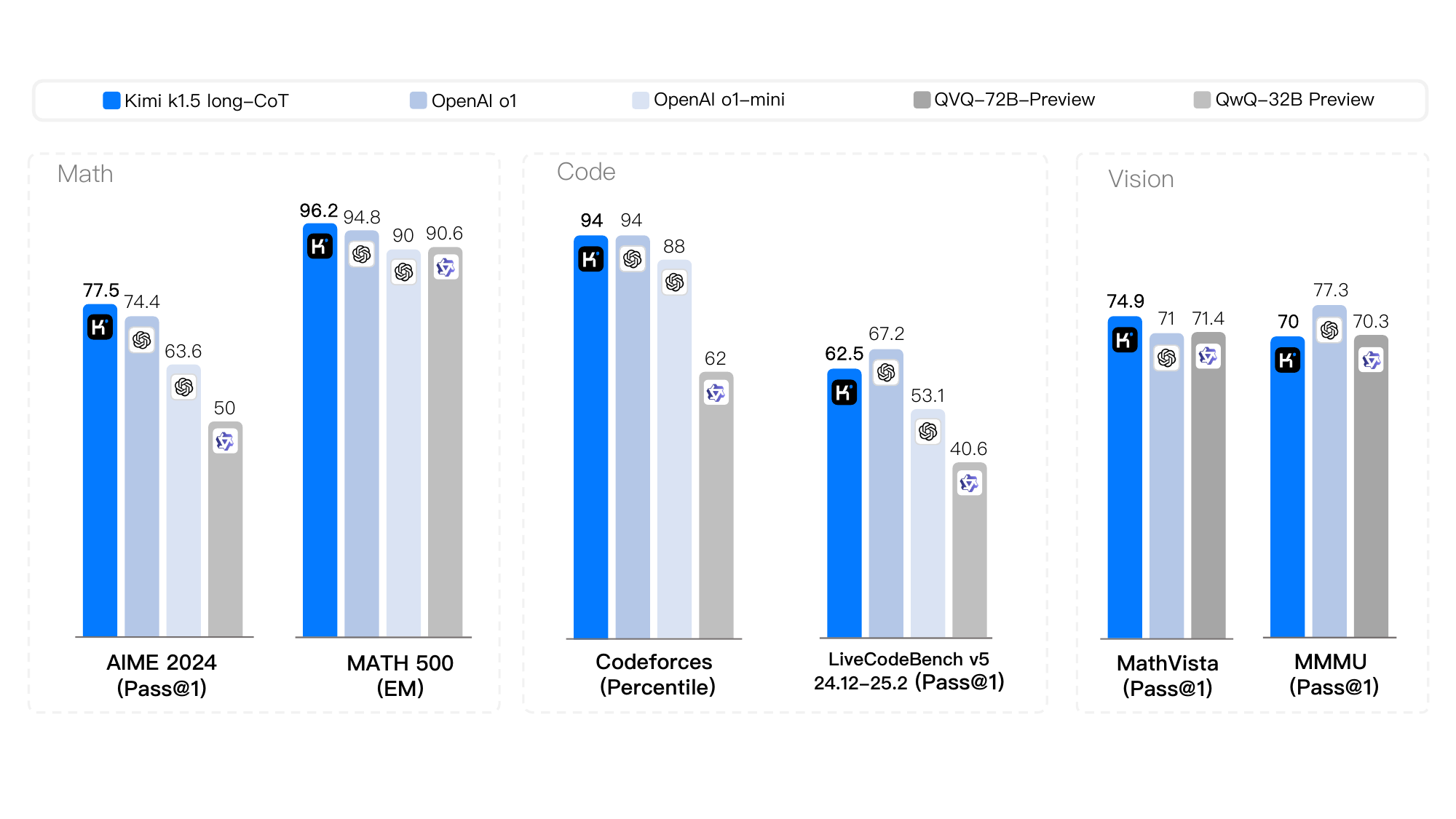}
    \caption{Kimi k1.5 long-CoT results.}
    \label{fig:long-COT-results}
\end{figure}

\begin{figure}[htb]
    \centering
    \vspace{-24pt}
    \adjustbox{center}{%
        \includegraphics[width=1.03\textwidth]{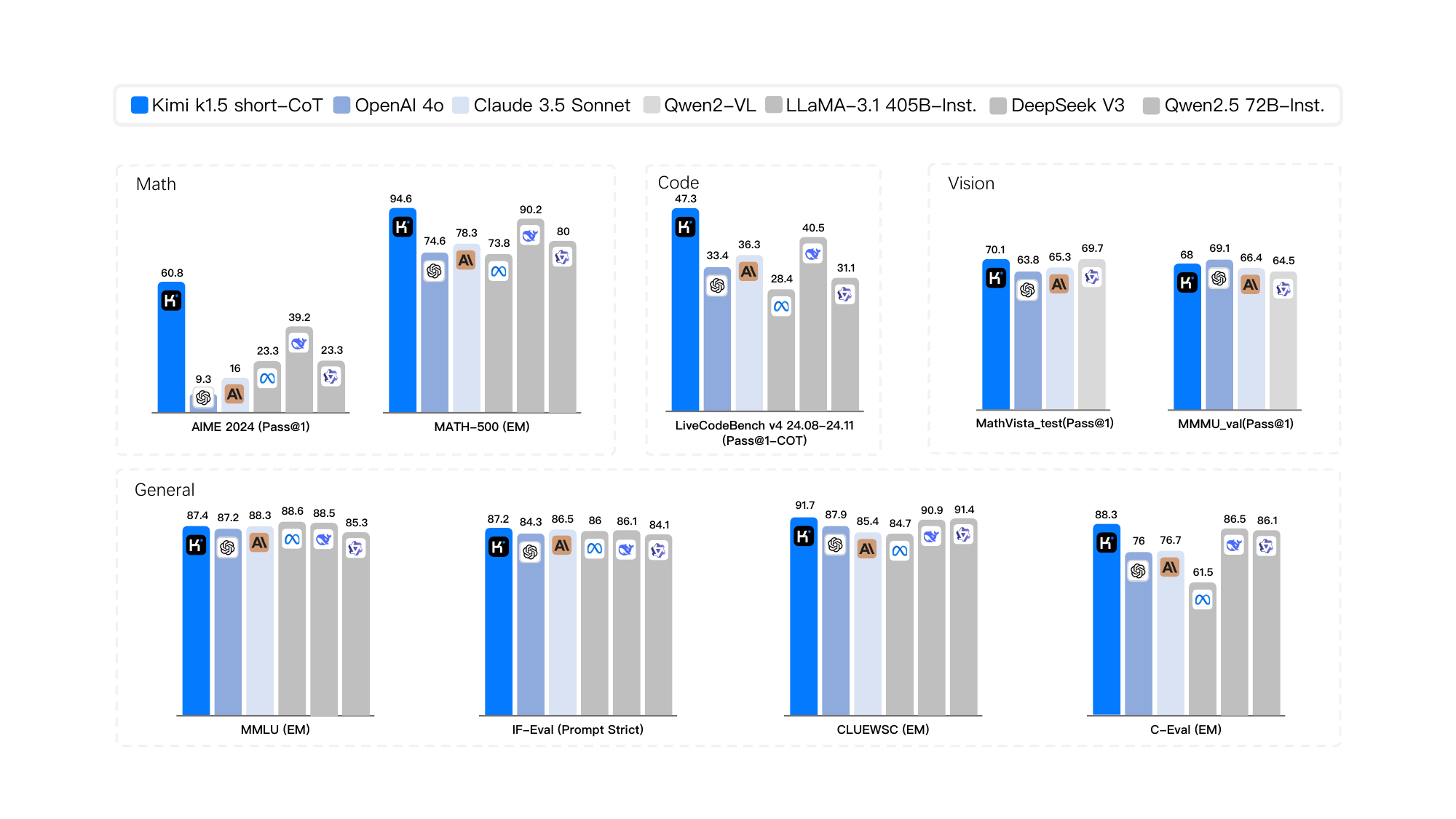}
    }
    \vspace{-32pt}
    \caption{Kimi k1.5 short-CoT results.}
    \label{fig:short-COT-results}
\end{figure}

\section{Introduction}

Language model pretraining with next token prediction has been studied under the context of the scaling law, where proportionally scaling model parameters and data sizes leads to the continued improvement of intelligence. ~\citep{kaplan2020scalinglawsneurallanguage,hoffmann2022trainingcomputeoptimallargelanguage} However, this approach is limited to the amount of available high-quality training data~\citep{villalobos2024rundatalimitsllm,muennighoff2023scalingdataconstrainedlanguagemodels}. In this report, we present the training recipe of Kimi k1.5, our latest multi-modal LLM trained with reinforcement learning (RL). The goal is to explore a possible new axis for continued scaling. Using RL with LLMs, the models learns to explore with rewards and thus is not limited to a pre-existing static dataset.

There are a few key ingredients about the design and training of k1.5.
\begin{itemize}
    \item \textbf{Long context scaling}. We scale the context window of RL to 128k and observe continued improvement of performance with an increased context length. A key idea behind our approach is to use partial rollouts to improve training efficiency---i.e., sampling new trajectories by reusing a large chunk of previous trajectories, avoiding the cost to re-generate the new trajectories from scratch. Our observation identifies the context length as a key dimension of the continued scaling of RL with LLMs.
    \item \textbf{Improved policy optimization}. We derive a formulation of RL with long-CoT and employ a variant of online mirror descent for robust policy optimization. This algorithm is further improved by our effective sampling strategy, length penalty, and optimization of the data recipe.
    \item \textbf{Simplistic Framework}. Long context scaling, combined with the improved policy optimization methods, establishes a simplistic RL framework for learning with LLMs. Since we are able to scale the context length, the learned CoTs exhibit the properties of planning, reflection, and correction. An increased context length has an effect of increasing the number of search steps. As a result, we show that strong performance can be achieved without relying on more complex techniques such as Monte Carlo tree search, value functions, and process reward models.
    \item \textbf{Multimodalities}. Our model is jointly trained on text and vision data, which has the capabilities of jointly reasoning over the two modalities.
\end{itemize}

Moreover, we present effective long2short methods that use long-CoT techniques to improve short-CoT models. Specifically, our approaches include applying length penalty with long-CoT activations and model merging.

Our long-CoT version achieves state-of-the-art reasoning performance across multiple benchmarks and modalities---e.g., 77.5 on AIME, 96.2 on MATH 500, 94-th percentile on Codeforces, 74.9 on MathVista---matching OpenAI's o1.
Our model also achieves state-of-the-art short-CoT reasoning results---e.g., 60.8 on AIME, 94.6 on MATH500, 47.3 on LiveCodeBench---outperforming existing short-CoT models such as GPT-4o and Claude Sonnet 3.5 by a large margin (up to +550\%). Results are shown in Figures \ref{fig:long-COT-results} and \ref{fig:short-COT-results}.

\section{Approach: Reinforcement Learning with LLMs}
\label{sec:post_training}

The development of Kimi k1.5 consists of several stages: pretraining, vanilla supervised fine-tuning (SFT), long-CoT supervised fine-turning, and reinforcement learning (RL).
This report focuses on RL, beginning with an overview of the RL prompt set curation (Section \ref{sec:rl_prompt}) and long-CoT supervised finetuning (Section \ref{sec:long_cot_sft}), followed by an in-depth discussion of RL training strategies in Section \ref{sec:reinforcement_learning}. 
Additional details on pretraining and vanilla supervised finetuning can be found in Section \ref{sec:others}.

\subsection{RL Prompt Set Curation} 
\label{sec:rl_prompt}
Through our preliminary experiments, we found that the quality and diversity of the RL prompt set play a critical role in ensuring the effectiveness of reinforcement learning. A well-constructed prompt set not only guides the model toward robust reasoning but also mitigates the risk of reward hacking and overfitting to superficial patterns. Specifically, three key properties define a high-quality RL prompt set:
\begin{itemize}
    \item \textbf{Diverse Coverage}: Prompts should span a wide array of disciplines, such as STEM, coding, and general reasoning, to enhance the model’s adaptability and ensure broad applicability across different domains.
    \item \textbf{Balanced Difficulty}: The prompt set should include a well-distributed range of easy, moderate, and difficult questions to facilitate gradual learning and prevent overfitting to specific complexity levels.
    \item \textbf{Accurate Evaluability}: Prompts should allow objective and reliable assessment by verifiers, ensuring that model performance is measured based on correct reasoning rather than superficial patterns or random guess. 
\end{itemize}

To achieve diverse coverage in the prompt set, we employ automatic filters to select questions that require rich reasoning and are straightforward to evaluate.
Our dataset includes problems from various domains, such as STEM fields, competitions, and general reasoning tasks, incorporating both text-only and image-text question-answering data.
Furthermore, we developed a tagging system to categorize prompts by domain and discipline, ensuring balanced representation across different subject areas~\citep{li2023tagging,liu2023tagging}.

We adopt a model-based approach that leverages the model’s own capacity to adaptively assess the difficulty of each prompt.  
Specifically, for every prompt, an SFT model generates answers ten times using a relatively high sampling temperature. The pass rate is then calculated and used as a proxy for the prompt's difficulty—the lower the pass rate, the higher the difficulty. This approach allows difficulty evaluation to be aligned with the model’s intrinsic capabilities, making it highly effective for RL training. By leveraging this method, we can prefilter most trivial cases and easily explore different sampling strategies during RL training.

To avoid potential reward hacking~\citep{everitt2021rewardtamperingproblemssolutions,pan2022the}, we need to ensure that both the reasoning process and the final answer of each prompt can be accurately verified. 
Empirical observations reveal that some complex reasoning problems may have relatively simple and easily guessable answers, leading to false positive verification—where the model reaches the correct answer through an incorrect reasoning process.
To address this issue, we exclude questions that are prone to such errors, such as multiple-choice, true/false, and proof-based questions.
Furthermore, for general question-answering tasks, we propose a simple yet effective method to identify and remove easy-to-hack prompts.
Specifically, we prompt a model to guess potential answers without any CoT reasoning steps.
If the model predicts the correct answer within $N$ attempts, the prompt is considered too easy-to-hack and removed.
We found that setting $N=8$ can remove the majority easy-to-hack prompts. Developing more advanced verification models remains an open direction for future research.

\subsection{Long-CoT Supervised Fine-Tuning}
\label{sec:long_cot_sft}

With the refined RL prompt set, we employ prompt engineering to construct a small yet high-quality long-CoT warmup dataset, containing accurately verified reasoning paths for both text and image inputs. 
This approach resembles rejection sampling (RS) but focuses on generating long-CoT reasoning paths through prompt engineering. 
The resulting warmup dataset is designed to encapsulate key cognitive processes that are fundamental to human-like reasoning, such as \textbf{planning}, where the model systematically outlines steps before execution; \textbf{evaluation}, involving critical assessment of intermediate steps; \textbf{reflection}, enabling the model to reconsider and refine its approach; and \textbf{exploration}, encouraging consideration of alternative solutions.
By performing a lightweight SFT on this warm-up dataset, we effectively prime the model to internalize these reasoning strategies. As a result, the fine-tuned long-CoT model demonstrates improved capability in generating more detailed and logically coherent responses, which enhances its performance across diverse reasoning tasks.

\subsection{Reinforcement Learning} 
\label{sec:reinforcement_learning}
\subsubsection{Problem Setting}

Given a training dataset $\mathcal{D} =\{(x_i, y^*_i)\}_{i=1}^n$ of problems $x_i$ and corresponding ground truth answers $y^*_i$, our goal is to train a policy model $\pi_\theta$ to accurately solve test problems.  
In the context of complex reasoning, the mapping of problem $x$ to solution $y$ is non-trivial. 
To tackle this challenge, the \emph{chain of thought} (CoT) method proposes to use a sequence of intermediate  steps $z = (z_1, z_2, \dots, z_m)$ to bridge $x$ and $y$, where each $z_i$ is a coherent sequence of  tokens that acts as a significant intermediate step toward solving the problem \citep{wei2022chain}. 
When solving problem $x$, 
 thoughts $z_t\sim \pi_\theta(\cdot | x, z_1,\dots, z_{t-1})$ are auto-regressively sampled, followed by the final answer $y\sim \pi_\theta(\cdot | x, z_1,\dots, z_{m})$. 
We use $y, z\sim \pi_\theta$ to denote this sampling procedure.  Note that both the thoughts and final answer are sampled as a language sequence. 

To further enhance the model's reasoning capabilities, \emph{planning} algorithms are employed to explore various thought processes, generating improved CoT at inference time \citep{yao2024tree,wu2024inference,snell2024scaling}. 
The core insight of these approaches is the explicit construction of a search tree of thoughts guided by value estimations. This allows the model to explore diverse continuations of a thought process or backtrack to investigate new directions when encountering dead ends.  
In more detail, let $\mathcal{T}$ be a search tree where each node represents a partial solution $s=(x,z_{1:|s|})$. Here $s$ consists of the problem $x$ and a sequence of thoughts $z_{1:|s|} = (z_1,\dots,z_{|s|})$ leading up to that node, with $|s|$ denoting number of thoughts in the sequence.  
The planning algorithm uses a critic model $v$ to provide feedback $v(x, z_{1:|s|})$, which helps evaluate the current progress towards solving the problem and identify any errors in the existing partial solution. 
We note that the feedback can be provided by either a discriminative score or a language sequence\citep{zhang2408generative}. 
Guided by the feedbacks for all $s\in\mathcal{T}$, the planning algorithm selects the most promising node for expansion, thereby growing the search tree. The above process repeats iteratively until a full solution is derived. 

We can also approach planning algorithms from an \emph{algorithmic perspective}.
Given past search history  available at the $t$-th iteration $(s_1, v(s_1),\dots,s_{t-1}, v(s_{t-1}))$, a planning algorithm $\mathcal{A}$ iteratively determines the next search direction $\mathcal{A}(s_{t} | s_1, v(s_1), \dots, s_{t-1}, v(s_{t-1}))$ and provides feedbacks for the current search progress $\mathcal{A}(v(s_{t}) | s_1, v(s_1), \dots, s_{t})$.  
Since both thoughts and feedbacks can be viewed as intermediate reasoning steps, 
and these components can both be represented as sequence of language tokens, we use $z$ to replace $s$ and $v$ to simplify the notations. 
Accordingly, we view a planning algorithm as a mapping that directly acts on a sequence of reasoning steps $\mathcal{A}(\cdot | z_1, z_2, \dots)$. In this framework, all information stored in the search tree used by the planning algorithm is flattened into the full context provided to the algorithm.
This provides an intriguing perspective on generating high-quality CoT: Rather than explicitly constructing a search tree and implementing a planning algorithm, we could potentially train a model to approximate this process. Here, the number of thoughts (i.e., language tokens) serves as an analogy to the computational budget traditionally allocated to planning algorithms. 
Recent advancements in long context windows facilitate seamless scalability during both the training and testing phases.
If feasible, this method enables the model to run an implicit search over the reasoning space directly via auto-regressive predictions. 
Consequently, the model not only learns to solve a set of training problems but also develops the ability to tackle individual problems effectively, leading to improved generalization to unseen test problems.

We thus consider training the model to generate CoT with reinforcement learning (RL) \citep{o12024}.   
Let $r$ be a reward model that justifies the correctness of the proposed answer $y$  for the given problem $x$ based on the ground truth $y^*$, by assigning a value $r(x, y, y^*)\in \{0,1\}$. 
For verifiable problems, the reward is directly determined by predefined criteria or rules. For example, in coding problems, we assess whether the answer passes the test cases. For problems with free-form ground truth, we train a reward model $r(x, y, y^*)$ that predicts if the answer matches the ground truth. 
Given a problem $x$, the model $\pi_\theta$ generates a CoT and the final answer through the sampling procedure $z\sim\pi_\theta(\cdot | x)$, $y\sim\pi_\theta(\cdot |x, z)$. 
The quality of the generated CoT is evaluated by whether it can lead to a correct final answer. 
In summary, we consider the following objective to optimize the policy
\begin{align}
\max_\theta \mathbb{E}_{(x, y^*)\sim\mathcal{D}, (y, z)\sim\pi_\theta}\left[ r(x, y, y^*) \right]\, .
\end{align}
By scaling up RL training, we aim to train a model that harnesses the strengths of both simple prompt-based CoT and planning-augmented CoT. 
The model still auto-regressively sample language sequence during inference, thereby circumventing the need for the complex parallelization required by advanced planning algorithms during deployment. 
However, a key distinction from simple prompt-based methods is that the model should not merely  follow a series of reasoning steps. Instead, it should also learn critical planning skills including error identification, backtracking and solution refinement by leveraging the entire set of explored thoughts as contextual information. 

\subsubsection{Policy Optimization}

We apply a variant of {online policy mirror decent} as our training algorithm \citep{abbasi2019politex,mei2019principled,tomar2020mirror}. 
The algorithm performs iteratively. 
At the $i$-th iteration, we use the current model $\pi_{\theta_i}$ as a reference model and optimize the following relative entropy regularized policy optimization problem,
\begin{align}
\max_\theta \mathbb{E}_{(x, y^*)\sim\mathcal{D}}\left[ \mathbb{E}_{(y, z)\sim\pi_\theta} \left[r(x, y, y^*)\right] - \tau \mathrm{KL} (\pi_{\theta}(x) || \pi_{\theta_i}(x))  \right]\, ,
\label{eq:opmd}
\end{align}
where $\tau > 0$ is a parameter controlling the degree of regularization. 
This objective has a closed form solution 
\begin{align*}
\pi^*(y, z | x) = \pi_{\theta_i}(y, z | x) \exp(r(x, y, y^*) / \tau) / Z\, .
\end{align*}
Here $Z=\sum_{y',z'} \pi_{\theta_i}(y', z' | x) \exp(r(x, y', y^*) / \tau)$ is the normalization factor. Taking logarithm of both sides we have for \emph{any} $(y, z)$ the following constraint is satisfied, which allows us to leverage off-policy data during optimization
\begin{align*}
r(x, y, y^*) - \tau \log Z = \tau \log \frac{\pi^*(y, z | x)} { \pi_{\theta_i}(y, z | x)}\, .
\end{align*}
This motivates the following surrogate loss
\begin{align*}
L(\theta) = \mathbb{E}_{(x,y^*)\sim\mathcal{D}}\left[ \mathbb{E}_{(y, z)\sim\pi_{\theta_i}} \left[  \left( r(x, y, y^*) - \tau \log Z - \tau \log \frac{\pi_\theta(y, z | x)}{{\pi}_{\theta_i}(y, z | x)} \right)^2 \right]\right] \, .
\end{align*}
To approximate $\tau\log Z$, we use samples $(y_1, z_1),\dots,(y_k,z_k)\sim \pi_{\theta_i}$: $\tau \log Z\approx \tau\log\frac{1}{k}\sum_{j=1}^k \exp(r(x, y_j, y^*)/\tau)$. 
We also find that using empirical mean of sampled rewards $\overline{r}=\mathrm{mean}(r(x,y_1,y^*),\dots,r(x, y_k, y^*)) $ yields effective practical results. 
This is reasonable since $\tau\log Z$ approaches the expected reward under $\pi_{\theta_i}$ as $\tau\rightarrow \infty$. 
Finally, we conclude our learning algorithm by taking the gradient of surrogate loss. 
For each problem $x$, $k$ responses are sampled using the reference policy $\pi_{\theta_i}$, 
and the gradient is given by
\begin{align}
\frac{1}{k} \sum_{j=1}^k \left( \nabla_\theta \log\pi_\theta (y_j, z_j | x) ( r(x, y_j, y^*) - \overline{r}) - \frac{\tau}{2} \nabla_{\theta} \left( \log \frac{\pi_\theta  (y_j, z_j | x) }{{\pi}_{\theta_i}  (y_j, z_j | x) } \right)^2 \right)\, .
\end{align}
To those familiar with policy gradient methods, this gradient resembles the policy gradient of \eqref{eq:opmd} using the mean of sampled rewards as the baseline \citep{kool2019buy,ahmadian2024back}. 
The main differences are  that the responses are sampled from $\pi_{\theta_i}$ rather than on-policy, and an $l_2$-regularization is applied. 
Thus we could see this as the natural extension of a usual on-policy regularized policy gradient algorithm to the off-policy case \citep{nachum2017bridging}. 
We sample a batch of problems from $\mathcal{D}$ and update the parameters to $\theta_{i+1}$, which subsequently serves as the reference policy for the next iteration. Since each iteration considers a different optimization problem due to the changing reference policy, we also reset the optimizer at the start of each iteration. 

We exclude the value network in our training system which has also been exploited in previous studies \citep{ahmadian2024back}. 
While this design choice significantly improves training efficiency, we also hypothesize that the conventional use of value functions for credit assignment in classical RL may not be suitable for our context. 
Consider a scenario where the model has generated a partial CoT $(z_1, z_2, \dots, z_t)$ and there are two potential next reasoning steps: $z_{t+1}$ and $z'_{t+1}$. 
Assume that $z_{t+1}$ directly leads to the correct answer, while $z'_{t+1}$ contains some errors. 
 If an oracle value function were accessible, it would indicate that $z_{t+1}$ preserves a higher value compared to $z'_{t+1}$. 
 According to the standard credit assignment principle, selecting $z'_{t+1}$ would be penalized as it has a negative advantages relative to the current policy. 
However, exploring $z'_{t+1}$ is extremely valuable for training the model to generate long CoT. 
By using the justification of the final answer derived from a long CoT as the reward signal, the model can learn the pattern of trial and error from taking $z'_{t+1}$ as long as it successfully recovers and reaches the correct answer.  
The key takeaway from this example is that we should encourage the model to explore diverse reasoning paths to enhance its capability in solving complex problems. This exploratory approach generates a wealth of experience that supports the development of critical planning skills. Our primary goal is not confined to attaining high accuracy on training problems but focuses on equipping the model with effective problem-solving strategies, ultimately improving its performance on test problems.

\subsubsection{Length Penalty}
\label{sec:lengthpenalty}
We observe an overthinking phenomenon that the model's response length significantly increases during RL training. 
Although this leads to better performance, an excessively lengthy reasoning process is costly during training and inference, and overthinking is often not preferred by humans.
To address this issue, we introduce a length reward to restrain the rapid growth of token length, thereby improving the model's token efficiency. 
Given $k$ sampled responses $(y_1, z_1),\dots,(y_k,z_k)$ of problem $x$ with true answer $y^*$, let $\mathrm{len}(i)$ be the length of $(y_i, z_i)$, $\mathrm{min\_len}=\min_i \mathrm{len}(i)$ and $\mathrm{max\_len}=\max_i \mathrm{len}(i)$. 
If $\mathrm{max\_len}=\mathrm{min\_len}$, we set length reward zero for all responses, as they have the same length. Otherwise the length reward is given by
\begin{align*}
\mathrm{len\_reward(i)} = 
\left\{
\begin{aligned}
\lambda & \quad \text{If}\ r(x, y_i, y^*) = 1 \\
\min(0, \lambda )& \quad \text{If}\ r(x, y_i, y^*) = 0\\
\end{aligned}
\right.\, ,\quad
\text{where } \lambda  = 0.5 - \frac{\mathrm{len}(i) - \mathrm{min\_len}}{\mathrm{max\_len} - \mathrm{min\_len}} \, .
\end{align*}
In essence, we promote shorter responses and penalize longer responses among correct ones, while explicitly penalizing long responses with incorrect answers. 
This length-based reward is then added to the original reward with a weighting parameter.

In our preliminary experiments, length penalty may slow down training during the initial phases. To alleviate this issue, we propose to gradually warm up the length penalty during training. Specifically, we employ standard policy optimization without length penalty, followed by a constant length penalty for the rest of training.

\subsubsection{Sampling Strategies}
\label{sec:sampling}

Although RL algorithms themselves have relatively good sampling properties (with more difficult problems providing larger gradients), their training efficiency is limited. Consequently, some well-defined prior sampling methods can yield potentially greater performance gains.
We exploit multiple signals to further improve the sampling strategy.
First, the RL training data we collect naturally come with different difficulty labels. For example, a math competition problem is more difficult than a primary school math problem.
Second, because the RL training process samples the same problem multiple times, we can also track the success rate for each individual problem as a metric of difficulty.
We propose two sampling methods to utilize these priors to improve training efficiency.

\paragraph{Curriculum Sampling}
We start by training on easier tasks and gradually progress to more challenging ones. Since the initial RL model has limited performance, spending a restricted computation budget on very hard problems often yields few correct samples, resulting in lower training efficiency. Meanwhile, our collected data naturally includes grade and difficulty labels, making difficulty-based sampling an intuitive and effective way to improve training efficiency.

\paragraph{Prioritized Sampling}
In addition to curriculum sampling, we use a prioritized sampling strategy to focus on problems where the model underperforms. We track the success rates \(s_i\) for each problem \(i\) and sample problems proportional to $1-s_i$, 
so that problems with lower success rates receive higher sampling probabilities. This directs the model’s efforts toward its weakest areas, leading to faster learning and better overall performance.

\subsubsection{More Details on Training Recipe}

\paragraph{Test Case Generation for Coding}

Since test cases are not available for many coding problems from the web, we design a method to automatically generate test cases that serve as a reward to train our model with RL. Our focus is primarily on problems that do not require a special judge. We also assume that ground truth solutions are available for these problems so that we can leverage the solutions to generate higher quality test cases.

We utilize the widely recognized test case generation library, CYaRon\footnote{https://github.com/luogu-dev/cyaron}, to enhance our approach. We employ our base Kimi k1.5 to generate test cases based on problem statements. The usage statement of CYaRon and the problem description are provided as the input to the generator. For each problem, we first use the generator to produce 50 test cases and also randomly sample 10 ground truth submissions for each test case. We run the test cases against the submissions. A test case is deemed valid if at least 7 out of 10 submissions yield matching results. After this round of filtering, we obtain a set of selected test cases. A problem and its associated selected test cases are added to our training set if at least 9 out of 10 submissions pass the entire set of selected test cases.

In terms of statistics, from a sample of 1,000 online contest problems, approximately 614 do not require a special judge. We developed 463 test case generators that produced at least 40 valid test cases, leading to the inclusion of 323 problems in our training set.

\paragraph{Reward Modeling for Math}
One challenge in evaluating math solutions is that different written forms can represent the same underlying answer. For instance, \(a^2 - 4\) and \((a+2)(a-2)\) may both be valid solutions to the same problem. We adopted two methods to improve the reward model’s scoring accuracy:

\begin{enumerate}
\item Classic RM:  
   Drawing inspiration from the InstructGPT \citep{ouyang2022training} methodology, we implemented a value-head based reward model and collected approximately 800k data points for fine-tuning. The model ultimately takes as input the “question,” the “reference answer,” and the “response,” and outputs a single scalar that indicates whether the response is correct.

\item Chain-of-Thought RM:  
   Recent research \citep{ankner2024critiqueoutloudrewardmodels, mcaleese2024llmcriticshelpcatch} suggests that reward models augmented with chain-of-thought (CoT) reasoning can significantly outperform classic approaches, particularly on tasks where nuanced correctness criteria matter—such as mathematics. Therefore, we collected an equally large dataset of about 800k CoT-labeled examples to fine-tune the Kimi model. Building on the same inputs as the Classic RM, the chain-of-thought approach explicitly generates a step-by-step reasoning process before providing a final correctness judgment in JSON format, enabling more robust and interpretable reward signals.
\end{enumerate}

During our manual spot checks, the Classic RM achieved an accuracy of approximately \textbf{84.4}, while the Chain-of-Thought RM reached \textbf{98.5} accuracy. In the RL training process, we adopted the Chain-of-Thought RM to ensure more correct feedback.

\paragraph{Vision Data}
To improve the model's real-world image reasoning capabilities and to achieve a more effective alignment between visual inputs and large language models (LLMs), our vision reinforcement learning (Vision RL) data is primarily sourced from three distinct categories: Real-world data, Synthetic visual reasoning data, and Text-rendered data.

\begin{enumerate}
\item The real-world data encompass a range of science questions across various grade levels that require graphical comprehension and reasoning, location guessing tasks that necessitate visual perception and inference, and data analysis that involves understanding complex charts, among other types of data. These datasets improve the model's ability to perform visual reasoning in real-world scenarios.

\item Synthetic visual reasoning data is artificially generated, including procedurally created images and scenes aimed at improving specific visual reasoning skills, such as understanding spatial relationships, geometric patterns, and object interactions. These synthetic datasets offer a controlled environment for testing the model’s visual reasoning capabilities and provide an endless supply of training examples.

\item Text-rendered data is created by converting textual content into visual format, enabling the model to maintain consistency when handling text-based queries across different modalities. By transforming text documents, code snippets, and structured data into images, we ensure the model provides consistent responses regardless of whether the input is pure text or text rendered as images (like screenshots or photos). This also helps to enhance the model's capability when dealing with text-heavy images.

\end{enumerate}

Each type of data is essential in building a comprehensive visual language model that can effectively manage a wide range of real-world applications while ensuring consistent performance across various input modalities.

\subsection{Long2short: Context Compression for Short-CoT Models}
\label{sec:long-to-short}

Though long-CoT models achieve strong performance, it consumes more test-time tokens compared to standard short-CoT LLMs. However, it is possible to transfer the thinking priors from long-CoT models to short-CoT models so that performance can be improved even with limited test-time token budgets. We present several approaches for this long2short problem, 
including model merging~\citep{yang2024model}, shortest rejection sampling, DPO~\citep{rafailov2024direct}, and long2short RL. 
Detailed descriptions of these methods are provided below:

\paragraph{Model Merging} Model merging has been found to be useful in maintaining generalization ability. We also discovered its effectiveness in improving token efficiency when merging a long-cot model and a short-cot model. This approach combines a long-cot model with a shorter model to obtain a new one without training. Specifically, we merge the two models by simply averaging their weights.
 \paragraph{Shortest Rejection Sampling} We observed that our model generates responses with a large length variation for the same problem. Based on this, we designed the Shortest Rejection Sampling method. This method samples the same question \(n\) times (in our experiments, \(n=8\)) and selects the shortest correct response for supervised fine-tuning.
 \paragraph{DPO} Similar with Shortest Rejection Sampling, we utilize the Long CoT model to generate multiple response samples. The shortest correct solution is selected as the positive sample, while longer responses are treated as negative samples, including both wrong longer responses and correct longer responses (1.5 times longer than the chosen positive sample). These positive-negative pairs form the pairwise preference data used for DPO training.
 \paragraph{Long2short RL} After a standard RL training phase, we select a model that offers the best balance between performance and token efficiency to serve as the base model, and conduct a separate long2short RL training phase. 
In this second phase, we apply the length penalty introduced in Section~\ref{sec:lengthpenalty}, and significantly reduce the maximum rollout length to further penalize responses that exceed the desired length while possibly correct.

\subsection{Other Training Details}
\label{sec:others}

\subsubsection{Pretraining}

The Kimi k1.5 base model is trained on a diverse, high-quality multimodal corpus. The language data covers five domains: English, Chinese, Code, Mathematics  Reasoning, and Knowledge. Multimodal data, including Captioning, Image-text Interleaving, OCR, Knowledge, and QA datasets, enables our model to acquire vision-language capabilities. Rigorous quality control ensures relevance, diversity, and balance in the overall pretrain dataset. Our pretraining proceeds in three stages: (1) Vision-language pretraining, where a strong language foundation is established, followed by gradual multimodal integration; (2) Cooldown, which consolidates capabilities using curated and synthetic data, particularly for reasoning and knowledge-based tasks; and (3) Long-context activation, extending sequence processing to 131,072 tokens. More details regarding our pretraining efforts can be found in Appendix~\ref{appendix:pretrain}.

\subsubsection{Vanilla Supervised Finetuning}
\label{sec:vanilla_supervised_finetuning}
We create the vanilla SFT corpus covering multiple domains. For non-reasoning tasks, including question-answering, writing, and text processing, we initially construct a seed dataset through human annotation. This seed dataset is used to train a seed model. Subsequently, we collect a diverse of prompts and employ the seed model to generate multiple responses to each prompt. Annotators then rank these responses and refine the top-ranked response to produce the final version. For reasoning tasks such as math and coding problems, where rule-based and reward modeling based verifications are more accurate and efficient than human judgment, we utilize rejection sampling to expand the SFT dataset.

Our vanilla SFT dataset comprises approximately 1 million text examples. Specifically, 500k examples are for general question answering, 200k for coding, 200k for math and science, 5k for creative writing, and 20k for long-context tasks such as summarization, doc-qa, translation, and writing. In addition, we construct 1 million text-vision examples encompassing various categories including chart interpretation, OCR, image-grounded conversations, visual coding, visual reasoning, and math/science problems with visual aids. 

We first train the model at the sequence length of 32k tokens for 1 epoch, followed by another epoch at the sequence length of 128k tokens. In the first stage (32k), the learning rate decays from $2\times10^{-5}$ to $2\times10^{-6}$, before it re-warmups to $1\times10^{-5}$ in the second stage~(128k) and finally decays to $1\times10^{-6}$. To improve training efficiency, we pack multiple training examples into each single training sequence.

\subsection{RL Infrastructure}
\label{sec:rl_infra}

\begin{figure}[htb]
    \centering
    \begin{subfigure}[b]{0.45\textwidth}
         \centering
         \includegraphics[width=\textwidth]{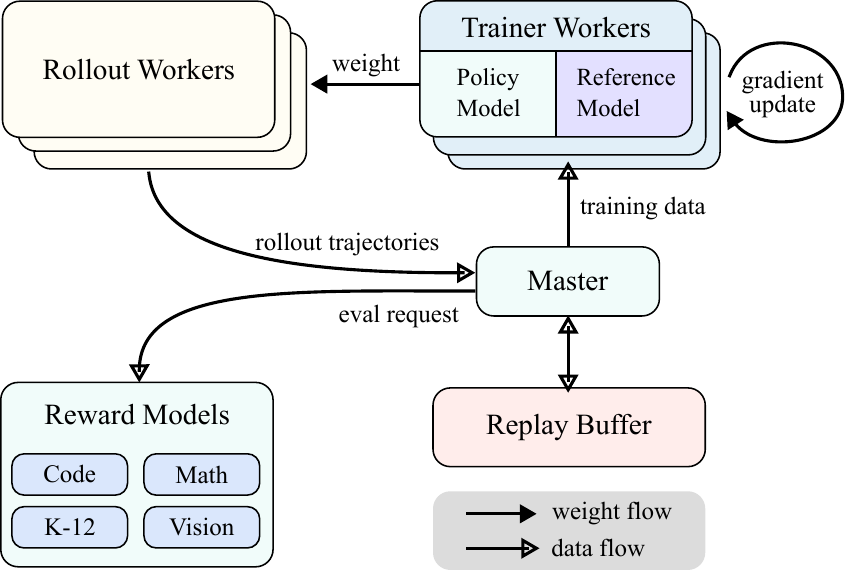}
         \caption{System overview}
         \label{fig:mshrl:overview}
    \end{subfigure}
    \hspace{0.05\textwidth}
    \begin{subfigure}[b]{0.3\textwidth}
         \centering
         \includegraphics[width=\textwidth]{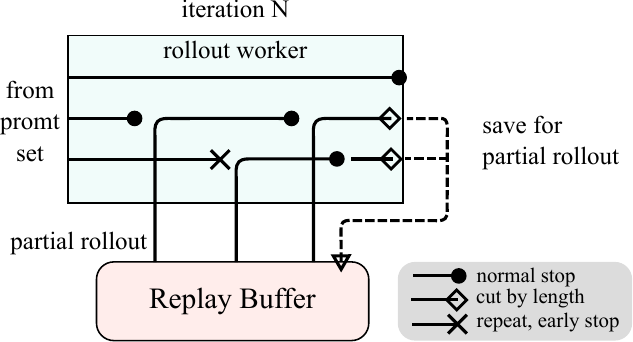}
         \caption{Partial Rollout}
         \label{fig:mshrl:replaybuffer}
    \end{subfigure}
    \caption{Large Scale Reinforcement Learning Training System for LLM}
    \label{fig:mshrl-framework}
\end{figure}

\subsubsection{Large Scale Reinforcement Learning Training System for LLM}

In the realm of artificial intelligence, reinforcement learning (RL) has emerged as a pivotal training methodology for large language models (LLMs)\citep{ouyang2022training}\citep{jaech2024openai}, drawing inspiration from its success in mastering complex games like Go, StarCraft II, and Dota 2 through systems such as AlphaGo\citep{silver2017mastering}, AlphaStar\citep{vinyals2019grandmaster}, and OpenAI Dota Five \citep{berner2019dota}. Following in this tradition, the Kimi k1.5 system adopts an iterative synchronous RL framework, meticulously designed to bolster the model's reasoning capabilities through persistent learning and adaptation. A key innovation in this system is the introduction of a Partial Rollout technique, designed to optimize the handling of complex reasoning trajectories.

The RL training system as illustrated in Figure~\ref{fig:mshrl:overview} operates through an iterative synchronous approach, with each iteration encompassing a rollout phase and a training phase. During the rollout phase, rollout workers, coordinated by a central master, generate rollout trajectories by interacting with the model, producing sequences of responses to various inputs. These trajectories are then stored in a replay buffer, which ensures a diverse and unbiased dataset for training by disrupting temporal correlations. In the subsequent training phase, trainer workers access these experiences to update the model's weights. This cyclical process allows the model to continuously learn from its actions, adjusting its strategies over time to enhance performance.

The central master serves as the central conductor, managing the flow of data and communication between the rollout workers, trainer workers, evaluation with reward models and the replay buffer. It ensures that the system operates harmoniously, balancing the load and facilitating efficient data processing.

The trainer workers access these rollout trajectories, whether completed in a single iteration or divided across multiple iterations, to compute gradient updates that refine the model's parameters and enhance its performance. This process is overseen by a reward model, which evaluates the quality of the model's outputs and provides essential feedback to guide the training process. The reward model's evaluations are particularly pivotal in determining the effectiveness of the model's strategies and steering the model towards optimal performance.

Moreover, the system incorporates a code execution service, which is specifically designed to handle code-related problems and is integral to the reward model. This service evaluates the model's outputs in practical coding scenarios, ensuring that the model's learning is closely aligned with real-world programming challenges. By validating the model's solutions against actual code executions, this feedback loop becomes essential for refining the model's strategies and enhancing its performance in code-related tasks.

\subsubsection{Partial Rollouts for Long CoT RL}
One of the primary ideas of our work is to scale long-context RL training. Partial rollouts is a key technique that effectively addresses the challenge of handling long-CoT features by managing the rollouts of both long and short trajectories. This technique establishes a fixed output token budget, capping the length of each rollout trajectory. If a trajectory exceeds the token limit during the rollout phase, the unfinished portion is saved to the replay buffer and continued in the next iteration. It ensures that no single lengthy trajectory monopolizes the system's resources. Moreover, since the rollout workers operate asynchronously, when some are engaged with long trajectories, others can independently process new, shorter rollout tasks. The asynchronous operation maximizes computational efficiency by ensuring that all rollout workers are actively contributing to the training process, thereby optimizing the overall performance of the system.

As illustrated in Figure~\ref{fig:mshrl:replaybuffer}, the partial rollout system works by breaking down long responses into segments across iterations (from iter n-m to iter n). The Replay Buffer acts as a central storage mechanism that maintains these response segments, where only the current iteration (iter n) requires on-policy computation. Previous segments (iter n-m to n-1) can be efficiently reused from the buffer, eliminating the need for repeated rollouts. This segmented approach significantly reduces the computational overhead: instead of rolling out the entire response at once, the system processes and stores segments incrementally, allowing for the generation of much longer responses while maintaining fast iteration times. During training, certain segments can be excluded from loss computation to further optimize the learning process, making the entire system both efficient and scalable.

The implementation of partial rollouts also offers repeat detection. The system identifies repeated sequences in the generated content and terminates them early, reducing unnecessary computation while maintaining output quality. 
Detected repetitions can be assigned additional penalties, effectively discouraging redundant content generation in the prompt set.

\subsubsection{Hybrid Deployment of Training and Inference}

\begin{figure}[htb]
    \centering
    \includegraphics[width=0.7\textwidth]{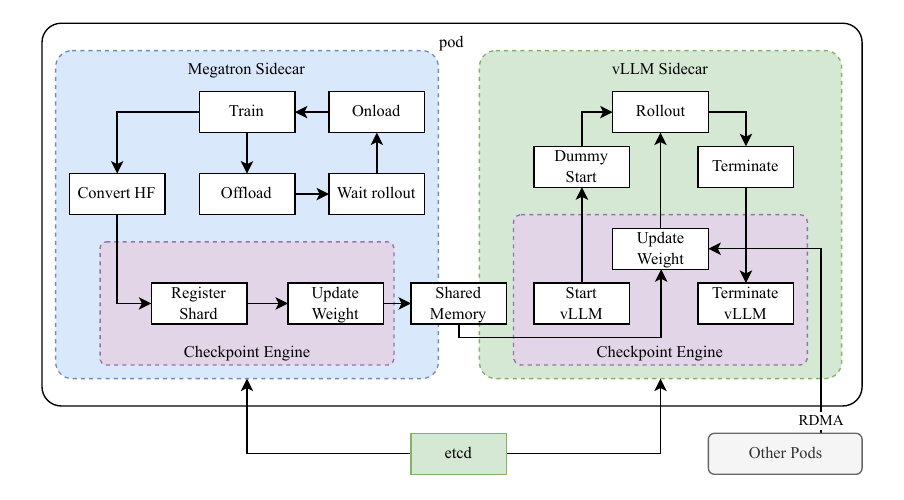}
    \caption{Hybrid Deployment Framework}
    \label{fig:infra_arch}
\end{figure}

The RL training process comprises of the following phases:

\begin{itemize}
    \item \textbf{Training Phase:} At the outset, Megatron \citep{shoeybi2020megatronlmtrainingmultibillionparameter} and vLLM \citep{kwon2023efficient} are executed within separate containers, encapsulated by a shim process known as checkpoint-engine (Section \ref{sec:checkpoint_engine}). Megatron commences the training procedure. After the training is completed, Megatron offloads the GPU memory and prepares to transfer current weights to vLLM.
    \item \textbf{Inference Phase:} Following Megatron's offloading, vLLM starts with dummy model weights and updates them with the latest ones transferred from Megatron via Mooncake \citep{qin2024mooncakekvcachecentricdisaggregatedarchitecture}. Upon completion of the rollout, the checkpoint-engine halts all vLLM processes.
    \item \textbf{Subsequent Training Phase:} Once the memory allocated to vLLM is released, Megatron onloads the memory and initiates another round of training.
\end{itemize}

We find existing works challenging to simultaneously support all the following characteristics.

\begin{itemize}
    \item Complex parallelism strategy: Megatron may have different parallelism strategy with vLLM. Training weights distributing in several nodes in Megatron could be challenging to be shared with vLLM. 
    \item Minimizing idle GPU resources: For On-Policy RL, recent works such as SGLang \citep{zheng2024sglangefficientexecutionstructured} and vLLM might reserve some GPUs during the training process, which conversely could lead to idle training GPUs. It would be more efficient to share the same devices between training and inference. 
    \item Capability of dynamic scaling: In some cases, a significant acceleration can be achieved by increasing the number of inference nodes while keeping the training process constant. Our system enables the efficient utilization of idle GPU nodes when needed. 
\end{itemize}

As illustrated in Figure~\ref{fig:infra_arch}, we implement this hybrid deployment framework (Section \ref{sec:hybrid_deployment}) on top of Megatron and vLLM, achieving less than one minute from training to inference phase and about ten seconds conversely. 

\paragraph{Hybrid Deployment Strategy}
\label{sec:hybrid_deployment}
We propose a hybrid deployment strategy for training and inference tasks, which leverages Kubernetes Sidecar containers sharing all available GPUs to collocate both workloads in one pod. The primary advantages of this strategy are:

\begin{itemize}
    \item It facilitates efficient resource sharing and management, preventing train nodes idling while waiting for inference nodes when both are deployed on separate nodes.
    \item Leveraging distinct deployed images, training and inference can each iterate independently for better performance. 
    \item The architecture is not limited to vLLM, other frameworks can be conveniently integrated.
\end{itemize}

\paragraph{Checkpoint Engine}
\label{sec:checkpoint_engine}
Checkpoint Engine is responsible for managing the lifecycle of the vLLM process, exposing HTTP APIs that enable triggering various operations on vLLM. For overall consistency and reliability, we utilize a global metadata system managed by the etcd service to broadcast operations and statuses.

It could be challenging to entirely release GPU memory by vLLM offloading primarily due to CUDA graphs, NCCL buffers and NVIDIA drivers. To minimize modifications to vLLM, we terminate and restart it when needed for better GPU utilization and fault tolerance. 

The worker in Megatron converts the owned checkpoints into the Hugging Face format in shared memory. This conversion also takes Pipeline Parallelism and Expert Parallelism into account so that only Tensor Parallelism remains in these checkpoints. Checkpoints in shared memory are subsequently divided into shards and registered in the global metadata system. We employ Mooncake to transfer checkpoints between peer nodes over RDMA. Some modifications to vLLM are needed to load weight files and perform tensor parallelism conversion.

\subsubsection{Code Sandbox}
\label{sec:code_box}

We developed the sandbox as a secure environment for executing user-submitted code, optimized for code execution and code benchmark evaluation. By dynamically switching container images, the sandbox supports different use cases through MultiPL-E~\citep{10103177}, DMOJ Judge Server \footnote{https://github.com/DMOJ/judge-server}, Lean, Jupyter Notebook, and other images.

For RL in coding tasks, the sandbox ensures the reliability of training data judgment by providing consistent and repeatable evaluation mechanisms. Its feedback system supports multi-stage assessments, such as code execution feedback and repo-level editing, while maintaining a uniform context to ensure fair and equitable benchmark comparisons across programming languages.

We deploy the service on Kubernetes for scalability and resilience, exposing it through HTTP endpoints for external integration. Kubernetes features like automatic restarts and rolling updates ensure availability and fault tolerance.

To optimize performance and support RL environments, we incorporate several techniques into the code execution service to enhance efficiency, speed, and reliability. These include:

\begin{itemize}
    \item \textbf{Using Crun:} We utilize \texttt{crun} as the container runtime instead of Docker, significantly reducing container startup times.
    \item \textbf{Cgroup Reusing:} We pre-create cgroups for container use, which is crucial in scenarios with high concurrency where creating and destroying cgroups for each container can become a bottleneck.
    \item \textbf{Disk Usage Optimization:} An overlay filesystem with an upper layer mounted as \texttt{tmpfs} is used to control disk writes, providing a fixed-size, high-speed storage space. This approach is beneficial for ephemeral workloads.
\end{itemize}

\begin{table}[h]
    \centering
    \begin{subtable}[t]{0.38\textwidth}
        \centering
        \vspace{-8pt} 
        \begin{tabular}{|c|c|}
            \hline
            \textbf{Method} & \textbf{Time (s)} \\
            \hline
            Docker & 0.12 \\
            Sandbox & 0.04 \\
            \hline
        \end{tabular}
        \caption{Container startup times}
        \vspace{-12pt}
        \label{tab:startup_times}
    \end{subtable}\hfill
    \begin{subtable}[t]{0.58\textwidth}
        \centering
        \vspace{-8pt}
        \begin{tabular}{|c|c|}
            \hline
            \textbf{Method} & \textbf{Containers/sec} \\
            \hline
            Docker & 27 \\
            Sandbox & 120 \\
            \hline
        \end{tabular}
        \caption{Maximum containers started per second on a 16-core machine}
        \vspace{-12pt}
        \label{tab:containers_per_second}
    \end{subtable}
\end{table}

These optimizations improve RL efficiency in code execution, providing a consistent and reliable environment for evaluating RL-generated code, essential for iterative training and model improvement.

\section{Experiments}

\subsection{Evaluation}

Since k1.5 is a multimodal model, we conducted comprehensive evaluation across various benchmarks for different modalities. The detailed evaluation setup can be found in Appendix \ref{sec:appendix_eval_detail}. Our benchmarks primarily consist of the following three categories:

\begin{itemize}
    \item \textbf{Text Benchmark}: MMLU~\citep{Hendrycks2020MeasuringMM}, IF-Eval~\citep{Zhou2023InstructionFollowingEF}, CLUEWSC~\citep{Xu2020CLUEAC}, C-EVAL~\citep{Huang2023CEvalAM} 
    \item \textbf{Reasoning Benchmark}:
    HumanEval-Mul, LiveCodeBench~\citep{Jain2024LiveCodeBenchHA}, Codeforces, AIME 2024, MATH-500~\citep{lightman2023lets}
    \item \textbf{Vision Benchmark}:
    MMMU~\citep{yue2024mmmu}, MATH-Vision~\citep{wang2024measuring}, MathVista~\citep{lu2023mathvista}
\end{itemize}

\subsection{Main Results}

\paragraph{K1.5 long-CoT model}

The performance of the Kimi k1.5 long-CoT model is presented in Table~\ref{tab:long_perf}. Through long-CoT supervised fine-tuning (described in Section~\ref{sec:long_cot_sft}) and vision-text joint reinforcement learning (discussed in Section~\ref{sec:reinforcement_learning}), the model's long-term reasoning capabilities are enhanced significantly. The test-time computation scaling further strengthens its performance, enabling the model to achieve state-of-the-art results across a range of modalities. Our evaluation reveals marked improvements in the model's capacity to reason, comprehend, and synthesize information over extended contexts, representing a advancement in multi-modal AI capabilities.

\paragraph{K1.5 short-CoT model}

The performance of the Kimi k1.5 short-CoT model is presented in Table~\ref{tab:short_perf}. This model integrates several techniques, including traditional supervised fine-tuning (discussed in Section~\ref{sec:vanilla_supervised_finetuning}), reinforcement learning (explored in Section~\ref{sec:reinforcement_learning}), and long-to-short distillation (outlined in Section~\ref{sec:long-to-short}). The results demonstrate that the k1.5 short-CoT model delivers competitive or superior performance compared to leading open-source and proprietary models across multiple tasks. These include text, vision, and reasoning challenges, with notable strengths in natural language understanding, mathematics, coding, and logical reasoning.

\begin{table}[h]
    \centering
    \begin{tabular}{@{}c l | c c | c c c@{}}
    \toprule
    & \multirow{3}{*}{\centering \textbf{Benchmark {\tiny (Metric)}}} 
    & \multicolumn{2}{c|}{\textbf{Language-only Model}} 
    & \multicolumn{3}{c}{\textbf{Vision-Language Model}} \\ 
    & & \textbf{QwQ-32B} & \textbf{OpenAI} & \textbf{QVQ-72B} & \textbf{OpenAI} & \textbf{Kimi} \\
    & & \textbf{Preview} & \textbf{o1-mini} & \textbf{Preview} & \textbf{o1} & \textbf{k1.5}  \\
    \midrule

    \multirow{4}{*}{Reasoning}
    & MATH-500 {\tiny (EM)} & 90.6 & 90.0 & - & 94.8 & \textbf{96.2} \\
    & AIME 2024 {\tiny (Pass@1)} & 50.0 & 63.6 & - & 74.4 & \textbf{77.5} \\
    & Codeforces {\tiny (Percentile)} & 62 & 88 & - & \textbf{94} & \textbf{94} \\
    & LiveCodeBench  {\tiny (Pass@1)} & 40.6 & 53.1 & - & \textbf{67.2} & 62.5 \\
    \midrule

    \multirow{3}{*}{Vision} 
    & MathVista-Test {\tiny (Pass@1)} & - & - & 71.4 & 71.0 & \textbf{74.9} \\
    & MMMU-Val {\tiny (Pass@1)} & - & - & 70.3 & \textbf{77.3} & 70.0 \\
    & MathVision-Full {\tiny (Pass@1)} & - & - & 35.9 & - & \textbf{38.6} \\

    \bottomrule
    \end{tabular}
    \vspace{1em}
    \caption{Performance of Kimi k1.5 long-CoT and  flagship open-source and proprietary models.}
    \label{tab:long_perf}
\end{table}

\begin{table}[h]
    \centering
    \footnotesize
    \setlength{\tabcolsep}{2.5pt}
    \begin{tabular}{@{}c l | c c c | c c c  c@{}}
    \toprule
    & \multirow{3}{*}{\centering \textbf{Benchmark {\tiny (Metric)}}} 
    & \multicolumn{3}{c|}{\textbf{Language-only Model}} 
    & \multicolumn{4}{c}{\textbf{Vision-Language Model}} \\ 
    & & \textbf{Qwen2.5} & \textbf{LLaMA-3.1} & \textbf{DeepSeek} & \textbf{Qwen2-VL} & \textbf{Claude-3.5-} & \textbf{GPT-4o}  & \textbf{Kimi} \\
    & & \textbf{72B-Inst.} & \textbf{405B-Inst.} & \textbf{V3}  & & \textbf{Sonnet-1022} & \textbf{0513} & \textbf{k1.5} \\
    \midrule
    
    \multirow{4}{*}{Text}
    & MMLU {\tiny (EM)}   & 85.3 & \textbf{88.6} & 88.5 & - & 88.3 & 87.2 & 87.4 \\
    & IF-Eval {\tiny (Prompt Strict)}  & 84.1 & 86.0 & 86.1 & - & 86.5 & 84.3 & \textbf{87.2} \\
    & CLUEWSC  {\tiny (EM)} & 91.4 & 84.7 & 90.9 & - & 85.4 & 87.9 & \textbf{91.7} \\
    & C-Eval  {\tiny (EM)}  & 86.1 & 61.5 & 86.5 & - & 76.7 & 76.0 & \textbf{88.3} \\
    \midrule
    
    \multirow{4}{*}{Reasoning} 
    & MATH-500 {\tiny (EM)} & 80.0 & 73.8 & 90.2 & - & 78.3 & 74.6 & \textbf{94.6} \\
    & AIME 2024 {\tiny (Pass@1)} & 23.3 & 23.3 & 39.2 & - & 16.0 & 9.3 & \textbf{60.8} \\
    & HumanEval-Mul {\tiny (Pass@1)}  & 77.3 & 77.2 & \textbf{82.6} & - & 81.7 & 80.5 & 81.5 \\
    & LiveCodeBench {\tiny (Pass@1)} & 31.1 & 28.4 & 40.5 & - & 36.3 & 33.4 & \textbf{47.3} \\
    \midrule
    
    \multirow{3}{*}{Vision} 
    & MathVista-Test {\tiny (Pass@1)} & - & - & - & 69.7 & 65.3 & 63.8 & \textbf{70.1} \\
    & MMMU-Val {\tiny (Pass@1)} & - & - & - & 64.5 & 66.4 & \textbf{69.1} & 68.0 \\
    & MathVision-Full {\tiny (Pass@1)} & - & - & - & 26.6 & \textbf{35.6} & 30.4 & 31.0 \\
 
    \bottomrule
    \end{tabular}
    \vspace{1em}
    \caption{Performance of Kimi k1.5 short-CoT and flagship open-source and proprietary models. VLM model performance were obtained from the OpenCompass benchmark platform (https://opencompass.org.cn/).}
    \label{tab:short_perf}
\end{table}

\subsection{Long Context Scaling}

\begin{figure}[htb]
    \centering
    \includegraphics[width=1.0\textwidth]{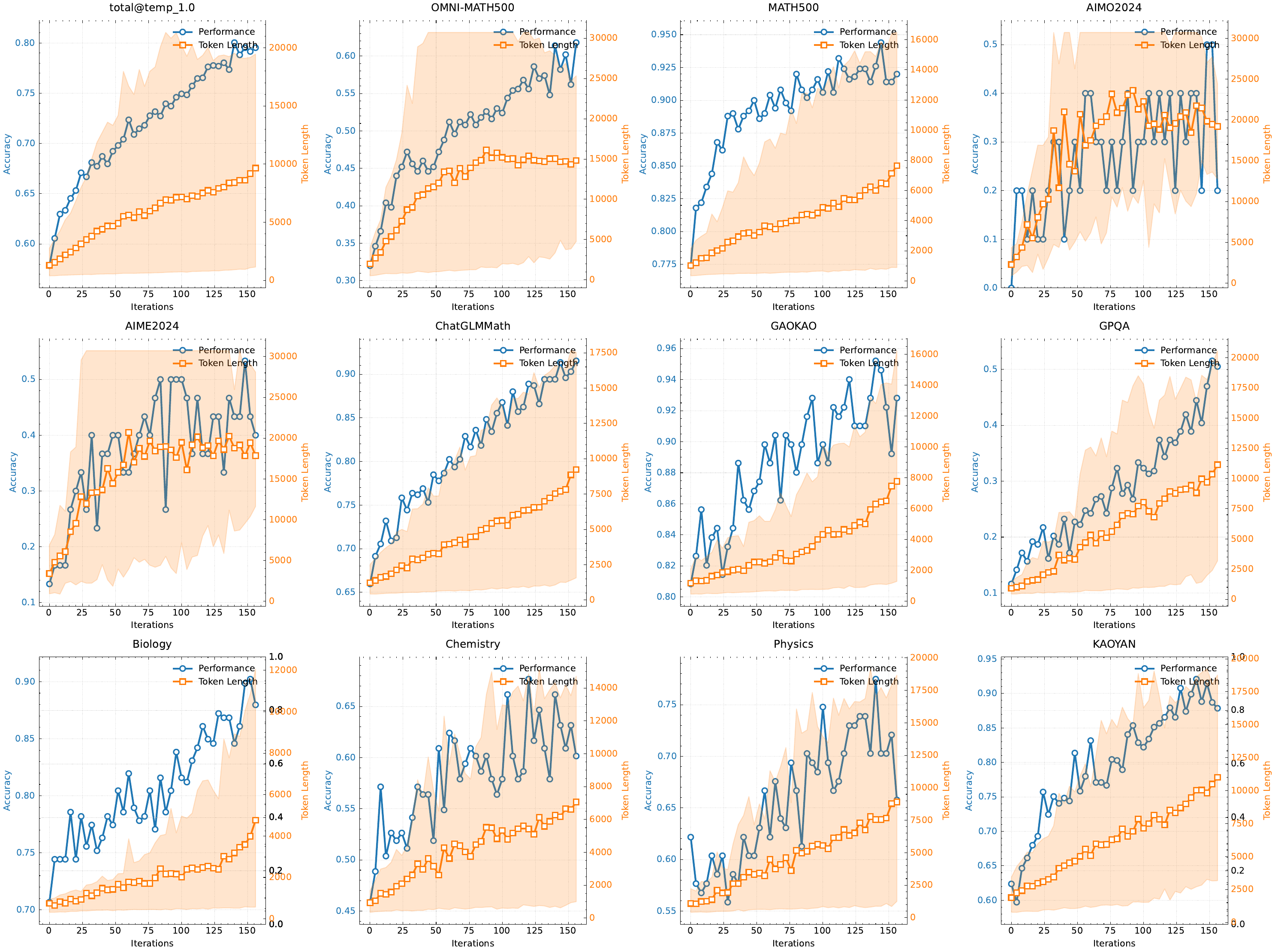}
    \caption{The changes on the training accuracy and length as train iterations grow. Note that the scores above come from an internal long-cot model with much smaller model size than k1.5 long-CoT model. 
    The shaded area represents the 95\% percentile of the response length.
    }
    \label{fig:perlen_vs_step}
\end{figure}

\begin{figure}[htb]
    \centering
    \includegraphics[width=1.0\textwidth]{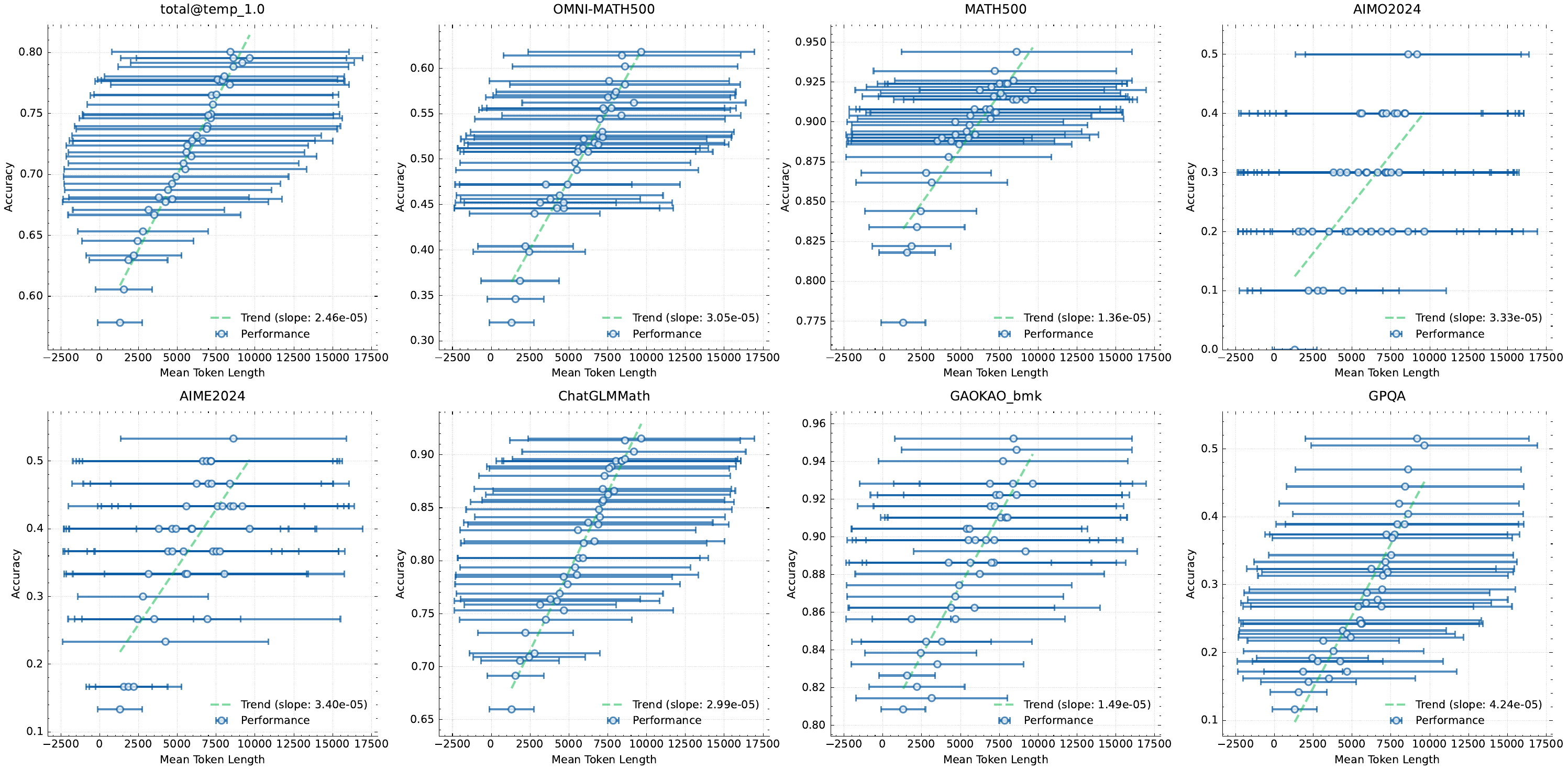}
    \caption{Model Performance Increases with Response Length}
    \label{fig:lenp}
\end{figure}

We employ a mid-sized model to study the scaling properties of RL with LLMs.
Figure \ref{fig:perlen_vs_step} illustrates the evolution of both training accuracy and response length across training iterations for the small model variant trained on the mathematical prompt set. As training progresses, we observe a concurrent increase in both response length and performance accuracy. Notably, more challenging benchmarks exhibit a steeper increase in response length, suggesting that the model learns to generate more elaborate solutions for complex problems. Figure \ref{fig:lenp} indicates a strong correlation between the model's output context length and its problem-solving capabilities.
Our final run of k1.5 scales to 128k context length and observes continued improvement on hard reasoning benchmarks. 

\begin{figure}[htb]
    \centering
    \includegraphics[width=1.0\textwidth]{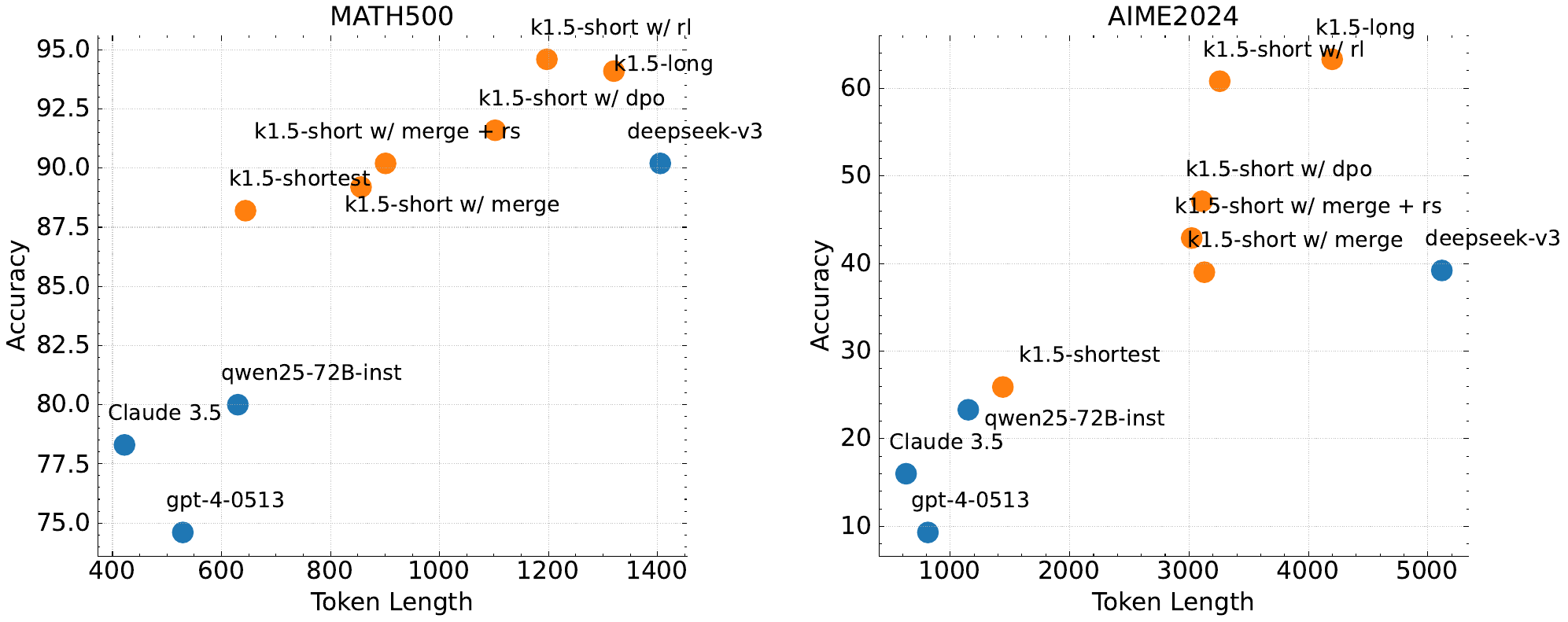}
    \caption{Long2Short Performance. All the k1.5 series  demonstrate better token efficiency compared to other models.}
    \label{fig:long2short}
\end{figure}

\subsection{Long2short}
We compared the proposed long2short RL algorithm with the DPO, shortest rejection sampling, and model merge methods introduced in the Section~\ref{sec:long-to-short}, focusing on the token efficiency for the long2short problem~\citep{chen2024not}, specifically how the obtained long-cot model can benefit a short model. In Figure~\ref{fig:long2short}, k1.5-long represents our long-cot model selected for long2short training. k1.5-short w/ rl refers to the short model obtained using the long2short RL training. k1.5-short w/ dpo denotes the short model with improved token efficiency through DPO training. k1.5-short w/ merge represents the model after model merging, while k1.5-short w/ merge + rs indicates the short model obtained by applying shortest rejection sampling to the merged model. k1.5-shortest represents the shortest model we obtained during the long2short training. As shown in Figure~\ref{fig:long2short}, the proposed long2short RL algorithm demonstrates the highest token efficiency compared other mehtods such as DPO and model merge. Notably, all models in the k1.5 series (marked in orange) demonstrate superior token efficiency compared to other models (marked in blue). For instance, k1.5-short w/ rl achieves a Pass@1 score of 60.8 on AIME2024 (averaged over 8 runs) while utilizing only 3,272 tokens on average. Similarly, k1.5-shortest attains a Pass@1 score of 88.2 on MATH500 while consuming approximately the same number of tokens as other short models.

\subsection{Ablation Studies}

\paragraph{Scaling of model size and context length}

\begin{figure}[htb]
    \centering
    \includegraphics[width=1.0\textwidth]{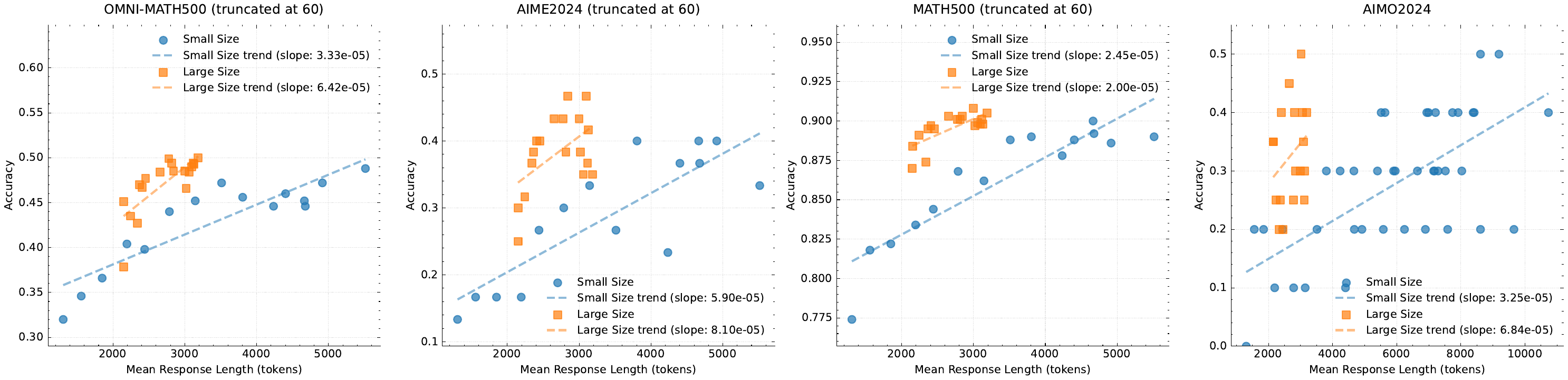}
    \caption{Model Performance vs Response Length of Different Model Sizes}
    \label{fig:length}
\end{figure}

Our main contribution is the application of RL to enhance the model's capacity for generating extended CoT, thereby improving its reasoning ability. 
A natural question arises: how does this compare to simply increasing the model size? To demonstrate the effectiveness of our approach, we trained two models of different sizes using the same dataset and recorded the evaluation results and average inference lengths from all checkpoints during RL training. These results are shown in Figure~\ref{fig:length}. Notably, although the larger model initially outperforms the smaller one, the smaller model can achieve comparable performance by utilizing longer CoTs optimized through RL. However, the larger model generally shows better token efficiency than the smaller model. 
This also indicates that if one targets the best possible performance, scaling the context length of a larger model has a higher upper bound and is more token efficient. However, if test-time compute has a budget, training smaller models with a larger context length may be viable solutions.

\begin{figure}[htb]
    \centering
    \includegraphics[width=1.0\textwidth]{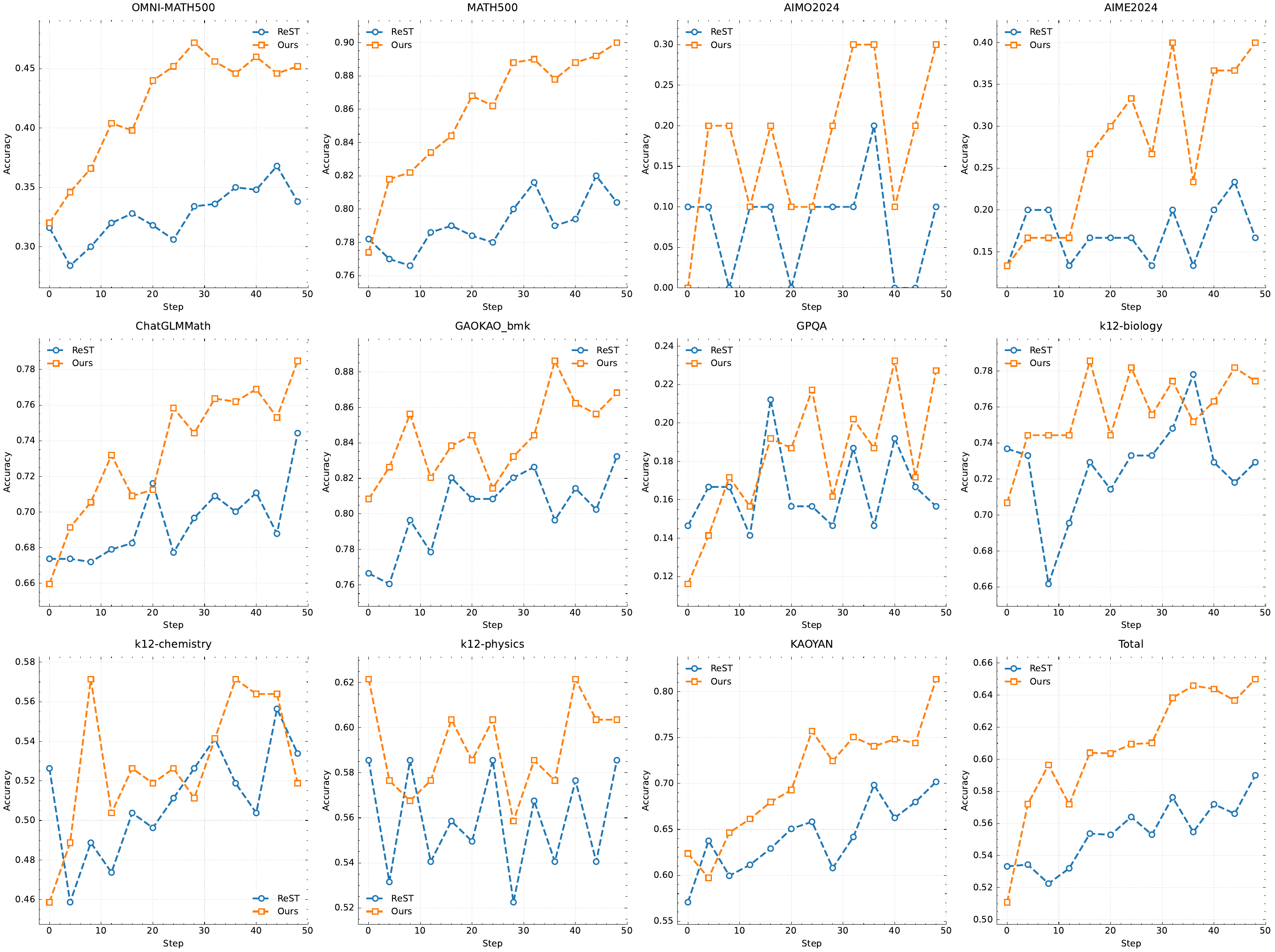}
    \caption{Comparison with using ReST for policy optimization. }
    \label{fig:rest}
\end{figure}

\paragraph{Effects of using negative gradients} 
We investigate the effectiveness of using ReST~\citep{gulcehre2023reinforced} as the policy optimization algorithm in our setting. 
The primary distinction between  ReST and other RL-based methods including ours is that ReST iteratively refines the model by fitting the best response sampled from the current model, without applying negative gradients to penalize incorrect responses. 
As illustrated in Figure~\ref{fig:rest}, our method exhibits superior sample complexity compared to ReST, indicating that the incorporation of negative gradients markedly enhances the model's efficiency in generating long CoT. Our method not only elevates the quality of reasoning but also optimizes the training process, achieving robust performance with fewer training samples. 
This finding suggests that the choice of policy optimization algorithm is crucial in our setting, as the performance gap between ReST and other RL-based methods is not as pronounced in other domains~\citep{gulcehre2023reinforced}. Therefore, our results highlight the importance of selecting an appropriate optimization strategy to maximize effectiveness in generating long CoT.

\paragraph{Sampling strategies}

\begin{figure}[htb]
    \centering
    \includegraphics[width=0.6\textwidth]{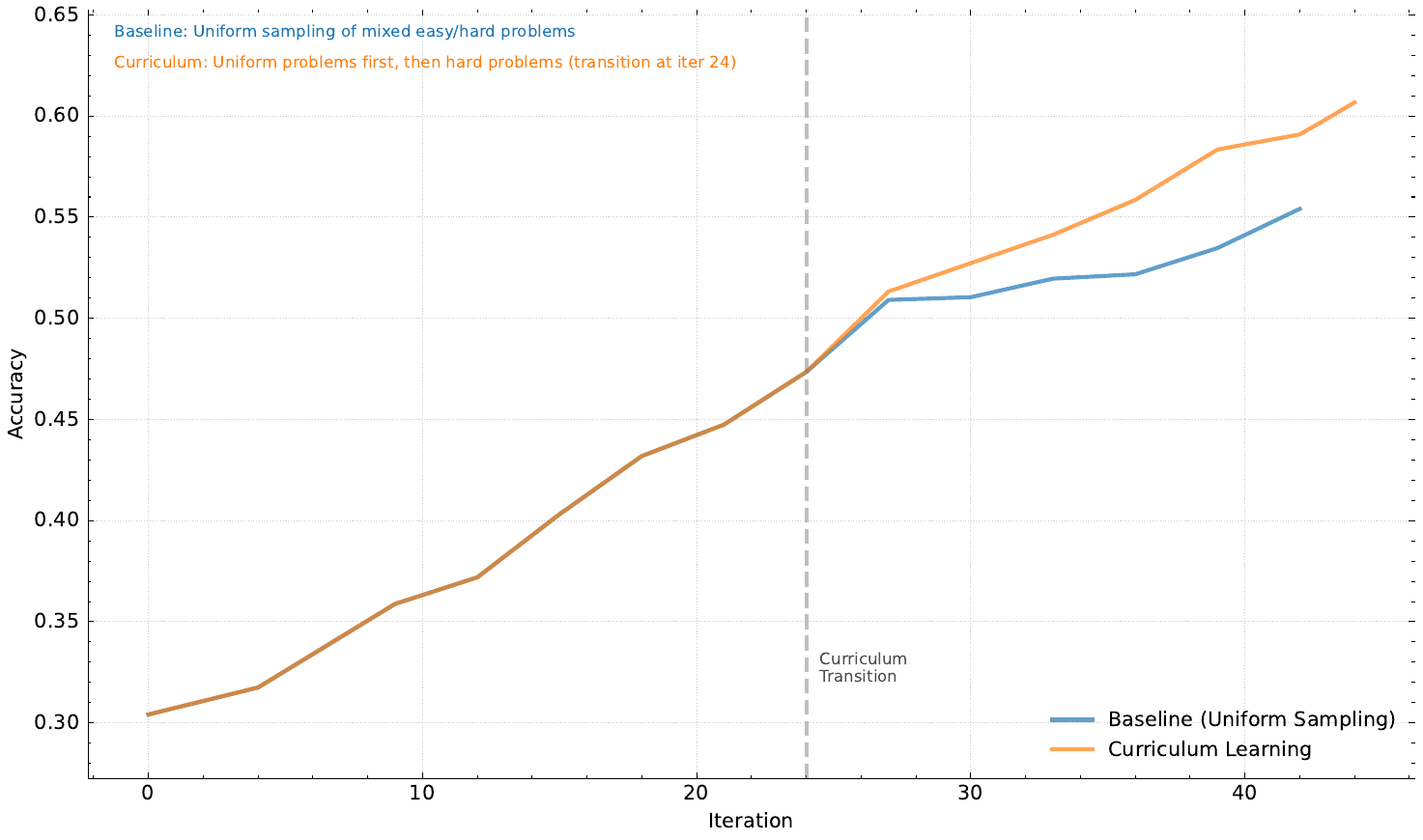}
    \caption{Analysis of curriculum learning approaches on model performance.}
    \label{fig:curriculum}
\end{figure}

We further demonstrate the effectiveness of our curriculum sampling strategy, as introduced in Section~\ref{sec:sampling}. Our training dataset $\mathcal{D}$ comprises a diverse mix of problems with varying levels of difficulty. 
With our curriculum sampling method, we initially use $\mathcal{D}$ for a warm-up phase and then focus solely on hard questions to train the model. 
This approach is compared to a baseline method that employs a uniform sampling strategy without any curriculum adjustments. As illustrated in Figure~\ref{fig:curriculum}, our results clearly show that the proposed curriculum sampling method significantly enhances the performance. 
This improvement can be attributed to the method's ability to progressively challenge the model, allowing it to develop a more robust understanding and competency in handling complex problems. By focusing training efforts on more difficult questions after an initial general introduction, the model can better strengthen its reasoning and problem solving capabilities.  

\section{Conclusions}

We present the training recipe and system design of k1.5, our latest multi-modal LLM trained with RL. One of the key insights we extract from our practice is that the scaling of context length is crucial to the continued improvement of LLMs. We employ optimized learning algorithms and infrastructure optimization such as partial rollouts to achieve efficient long-context RL training. How to further improve the efficiency and scalability of long-context RL training remains an important question moving forward.

Another contribution we made is a combination of techniques that enable improved policy optimization. Specifically, we formulate long-CoT RL with LLMs and derive a variant of online mirror descent for robust optimization. We also experiment with sampling strategies, length penalty, and optimizing the data recipe to achieve strong RL performance.

We show that strong performance can be achieved by long context scaling and improved policy optimization, even without using more complex techniques such as Monte Carlo tree search, value functions, and process reward models. In the future, it will also be intriguing to study improving credit assignments and reducing overthinking without hurting the model's exploration abilities.

We have also observed the potential of long2short methods. These methods largely improve performance of short CoT models. Moreover, it is possible to combine long2short methods with long-CoT RL in an iterative way to further increase token efficiency and extract the best performance out of a given context length budget. 

\printbibliography[title={References}]

\newpage
\appendix

\section*{Appendix}

\section{Contributions}

\begin{multicols}{2}
\noindent
\textbf{Research \& Development} \\

Angang Du \\
Bofei Gao \\
Bowei Xing \\
Changjiu Jiang \\
Cheng Chen \\
Cheng Li \\
Chenjun Xiao \\
Chenzhuang Du \\
Chonghua Liao* \\
Congcong Wang \\
Dehao Zhang \\
Enming Yuan \\
Enzhe Lu \\
Flood Sung \\
Guokun Lai \\
Haiqing Guo \\
Han Zhu \\
Hao Ding \\
Hao Hu \\
Hao Yang \\
Hao Zhang \\
Haotian Yao \\
Haotian Zhao \\
Haoyu Lu \\
Hongcheng Gao \\
Huan Yuan \\
Huabin Zheng \\
Jingyuan Liu \\
Jianlin Su \\
Jianzhou Wang \\
Jin Zhang \\
Junjie Yan \\
Lidong Shi \\
Longhui Yu \\
Mengnan Dong \\
Neo Zhang \\
Ningchen Ma* \\
Qiwei Pan \\
Qucheng Gong \\
Shaowei Liu \\
Shupeng Wei \\
Sihan Cao \\
Tao Jiang \\
Weimin Xiong \\
Weiran He \\
Weihao Gao* \\
Weixiao Huang \\
Weixin Xu \\
Wenhao Wu \\
Wenyang He \\
Xianqing Jia \\
Xingzhe Wu \\
Xinran Xu \\
Xinyu Zhou \\
Xinxing Zu \\
Xuehai Pan \\
Yang Li \\
Yangyang Hu \\
Yangyang Liu \\
Yanru Chen \\
Yejie Wang \\
Yidao Qin \\
Yibo Liu \\
Yiping Bao \\
Yifeng Liu* \\
Yulun Du \\
Yuzhi Wang \\
Yuxin Wu \\
Y. Charles \\
Zaida Zhou \\
Zhaoji Wang \\
Zhaowei Li \\
Zheng Zhang \\
Zhexu Wang \\
Zhiqi Huang \\
Zhilin Yang \\
Zihao Huang \\
Ziyao Xu \\
Zonghan Yang \\
Zongyu Lin \\

\noindent
\textbf{Data Annotation} \\

Chuning Tang \\
Fengxiang Tang  \\
Guangda Wei \\
Haoze Li \\
Haozhen Yu \\
Jia Chen \\
Jianhang Guo \\
Jie Zhao \\
Junyan Wu \\
Ling Ye \\
Shengling Ma \\
Siying Huang \\
Xianghui Wei \\
Yangyang Liu \\
Ying Yang \\
Zhen Zhu \\

\end{multicols}

The listing of authors is in alphabetical order based on their first names. Names marked with an asterisk (*) indicate people who are no longer part of our team.

\section{Pretraining}
\label{appendix:pretrain}

Reinforcement learning (RL) efficiency is closely tied to the performance of the underlying base model.
Frontier models such as Gemini\citep{geminiteam2024geminifamilyhighlycapable} and Llama\citep{grattafiori2024llama3herdmodels} highlight the importance of pretraining data quality in achieving high performance. 
However, many recent open-source models lack full transparency regarding their data processing pipelines and recipes, creating challenges for broader community understanding. While we are not open-sourcing our proprietary model at this time, we are committed to providing a comprehensive disclosure of our data pipeline and methodologies. In this section, we focus primarily on the multimodal pretraining data recipe, followed by a brief discussion of the model architecture and training stages.

\subsection{Language Data}

Our pretrain corpus is designed to provide comprehensive and high-quality data for training large language models (LLMs). It encompasses five domains: English, Chinese, Code, Mathematics \& Reasoning, and Knowledge. 
We employ sophisticated filtering and quality control mechanisms for each domain to ensure the highest quality training data. For all pretrain data, we conducted rigorous individual validation for each data source to assess its specific contribution to the overall training recipe. This systematic evaluation ensures the quality and effectiveness of our diverse data composition.

\paragraph{English and Chinese textual data} we developed a multi-dimensional quality filtering framework that combines multiple scoring methods to reduce individual biases and ensure comprehensive quality assessment. Our framework incorporates:\begin{enumerate}
   \item \textbf{Rule-based filtering}: We implement domain-specific heuristics to remove problematic content, including duplicate content, machine-translated text, and low-quality web scrapes. We also filter out documents with excessive special characters, unusual formatting, or spam patterns.
   
   \item \textbf{FastText-based classification}: We trained specialized FastText\citep{joulin2016bag,li2024datacomp} models to identify content quality based on linguistic features and semantic coherence. This helps identify documents with natural language flow and proper grammatical structure.

   \item \textbf{Embedding-based similarity analysis}: Using document embeddings \citep{chen2024bge}, we compute document-level similarity scores to identify and remove near-duplicates while preserving semantically valuable variations. This approach helps maintain diversity in our training corpus.
   
   \item \textbf{LLM-based quality assessment}: Following \citep{penedo2024fineweb}, we leverage LLMs to score documents based on coherence, informativeness, and potential educational value. This method is particularly effective at identifying nuanced quality indicators that simpler methods might miss.
\end{enumerate}
The final quality score for each document is calculated as a combination of these individual scores. Based on extensive empirical analysis, we implement dynamic sampling rates, where high-quality documents are upsampled, while low-quality documents are downsampled during training. 

\paragraph{Code data} The code data primarily consists of two categories. For the pure code data derived from code files, we adhered to the methodology of BigCode~\citep{li2023starcodersourceyou,lozhkov2024starcoder2stackv2} and conducted a comprehensive preprocessing of the dataset. Initially, we eliminated miscellaneous languages and applied a rule-based cleaning procedure to enhance data quality. Subsequently, we addressed language imbalance through strategic sampling techniques. Specifically, markup languages such as JSON, YAML, and YACC were down-sampled, while 32 major programming languages, including Python, C, C++, Java, and Go, were up-sampled to ensure a balanced representation. Regarding the text-code interleaved data sourced from various data sources, we use an embedding-based method to recall high-quality data. This approach ensures the diversity of the data and maintains its high quality.

\paragraph{Math \& Reasoning data} The mathematics and reasoning component of our dataset is crucial for developing strong analytical and problem-solving capabilities. 
The mathematical pre-training data are mainly retrieved from web text and PDF documents collected from publicly available internet sources.~\citep{paster2023openwebmath}
Initially, we discovered that our general-domain text extraction, data cleaning process and OCR models exhibited high false negative rates in the mathematical domain. Therefore, we first developed specialized data cleaning procedures and OCR models specifically for mathematical content, aiming to maximize the recall rate of mathematical data.
Subsequently, we implemented a two-stage data cleaning process: 
\begin{enumerate}
    \item Using FastText model for initial cleaning to remove most irrelevant data.
    \item Utilizing a fine-tuned language model to further clean the remaining data, resulting in high-quality mathematical data.
\end{enumerate}

\paragraph{Knowledge data} The knowledge corpus is meticulously curated to ensure a comprehensive coverage in academic disciplines. Our knowledge base primarily consists of academic exercises, textbooks, research papers, and other general educational literature. A significant portion of these materials is digitized through OCR processing, for which we have developed proprietary models optimized for academic content, particularly for handling mathematical formulas and special symbols.

We employ internal language models to annotate documents with multi-dimensional labels, including:
\begin{enumerate}
    \item OCR quality metrics to assess recognition accuracy
    \item Educational value indicators measuring pedagogical relevance
    \item Document type classification (e.g., exercises, theoretical materials)
\end{enumerate}

Based on these multi-dimensional annotations, we implement a sophisticated filtering and sampling pipeline. First and foremost, documents are filtered through OCR quality thresholds. Our OCR quality assessment framework places special attention on detecting and filtering out common OCR artifacts, particularly repetitive text patterns that often indicate recognition failures.

Beyond basic quality control, we carefully evaluate the educational value of each document through our scoring system. Documents with high pedagogical relevance and knowledge depth are prioritized, while maintaining a balance between theoretical depth and instructional clarity. This helps ensure that our training corpus contains high-quality educational content that can effectively contribute to the model's knowledge acquisition.

Finally, to optimize the overall composition of our training corpus, the sampling strategy for different document types is empirically determined through extensive experimentation. We conduct isolated evaluations to identify document subsets that contribute most significantly to the model's knowledge acquisition capabilities. These high-value subsets are upsampled in the final training corpus. However, to maintain data diversity and ensure model generalization, we carefully preserve a balanced representation of other document types at appropriate ratios. This data-driven approach helps us optimize the trade-off between focused knowledge acquisition and broad generalization capabilities.

\subsection{Multimodal Data}

Our multi-modal pretraining corpus is designed to provide high-quality data that enables models to process and understand information from multiple modalities, including text, images, and videos. 
To this end, we also have curated high-quality data from five categories—captioning, interleaving, OCR (Optical Character Recognition), knowledge, and general question answering—to form the corpus.

When constructing our training corpus, we developed several multi-modal data processing pipelines to ensure data quality, encompassing filtering, synthesis, and deduplication. 
Establishing an effective multi-modal data strategy is crucial during the joint training of vision and language, as it both preserves the capabilities of the language model and facilitates alignment of knowledge across diverse modalities.

We provide a detailed description of these sources in this section, which is organized into the following categories:

\paragraph{Caption data}
Our caption data provides the model with fundamental modality alignment and a broad range of world knowledge. By incorporating caption data, the multi-modal LLM gains wider world knowledge with high learning efficiency. We have integrated various open-source Chinese and English caption datasets like \citep{schuhmann2022laion, gadre2024datacomp} and also collected substantial in-house caption data from multiple sources. However, throughout the training process, we strictly limit the proportion of synthetic caption data to mitigate the risk of hallucination stemming from insufficient real-world knowledge.

For general caption data, we follow a rigorous quality control pipeline that avoids duplication and maintain high image-text correlation. We also vary image resolution during pretraining to ensure that the vision tower remains effective when processing images of both high- and low-resolution.

\textbf{Image-text interleaving data}
During the pretraining phase, model is benefit from interleaving data for many aspects, for example, multi-image comprehension ability can be boosted by interleaving data; interleaving data always provide detailed knowledge for the given image; a longer multi-modal context learning ability can also be gained by the interleaving data. What's more, we also find that interleaving data can contributes positively to maintaining the model’s language abilities. 
Thus, image-text interleaving data is an important part in our training corpus. 
Our multi-modal corpus considered open-sourced interleave datasets like ~\citep{zhu2024multimodal,laurenccon2024obelics} and also constructed large-scale in-house data using resources like textbooks, webpages and tutorials.
Further, we also find that synthesizing the interleaving data benefits the performance of multi-modal LLM for keeping the text knowledges.  
To ensure each image's knowledge is sufficiently studied, for all the interleaving data, other than the standard filtering, deduping and other quality control pipeline, we also integrated a data reordering procedure for keeping all the image and text in the correct order.  
 
\textbf{OCR data} 
Optical Character Recognition (OCR) is a widely adopted technique that converts text from images into an editable format. In k1.5, a robust OCR capability is deemed essential for better aligning the model with human values. Accordingly, our OCR data sources are diverse, ranging from open-source to in-house datasets, and encompassing both clean and augmented images.

In addition to the publicly available data, we have developed a substantial volume of in-house OCR datasets, covering multilingual text, dense text layouts, web-based content, and handwritten samples. 
Furthermore, following the principles outlined in OCR 2.0~\citep{wei2024general}, our model is also equipped to handle a variety of optical image types, including figures, tables, geometry diagrams, mermaid plots, and natural scene text. We apply extensive data augmentation techniques—such as rotation, distortion, color adjustments, and noise addition—to enhance the model’s robustness. As a result, our model achieves a high level of proficiency in OCR tasks.

\textbf{Knowledge data} The concept of multi-modal knowledge data is analogous to the previously mentioned text pretraining data, except here we focus on assembling a comprehensive repository of human knowledge from diverse sources to further enhance the model's capabilities. 
For example, carefully curated geometry data in our dataset is vital for developing visual reasoning skills, ensuring the model can interpret the abstract diagrams created by humans.

Our knowledge corpus adheres to a standardized taxonomy to balance content across various categories, ensuring diversity in data sources. Similar to text-only corpora, which gather knowledge from textbooks, research papers, and other academic materials, multi-modal knowledge data employs both a layout parser and an OCR model to process content from these sources. While we also include filtered data from internet-based and other external resources.

Because a significant portion of our knowledge corpus is sourced from internet-based materials, infographics can cause the model to focus solely on OCR-based information. In such cases, relying exclusively on a basic OCR pipeline may limit training effectiveness. To address this, we have developed an additional pipeline that better captures the purely textual information embedded within images.

\textbf{General QA Data}  
During the training process, we observed that incorporating a substantial volume of high-quality QA datasets into pretraining offers significant benefits. Specifically, we included rigorous academic datasets addressing tasks such as grounding, table/chart question answering, web agents, and general QA. In addition, we compiled a large amount of in-house QA data to further enhance the model’s capabilities. To maintain balanced difficulty and diversity, we applied scoring models and meticulous manual categorization to our general question answering dataset, resulting in overall performance improvements.

\subsection{Model Architecture}
\begin{figure}[t]
\centering
\includegraphics[width=0.9\textwidth]{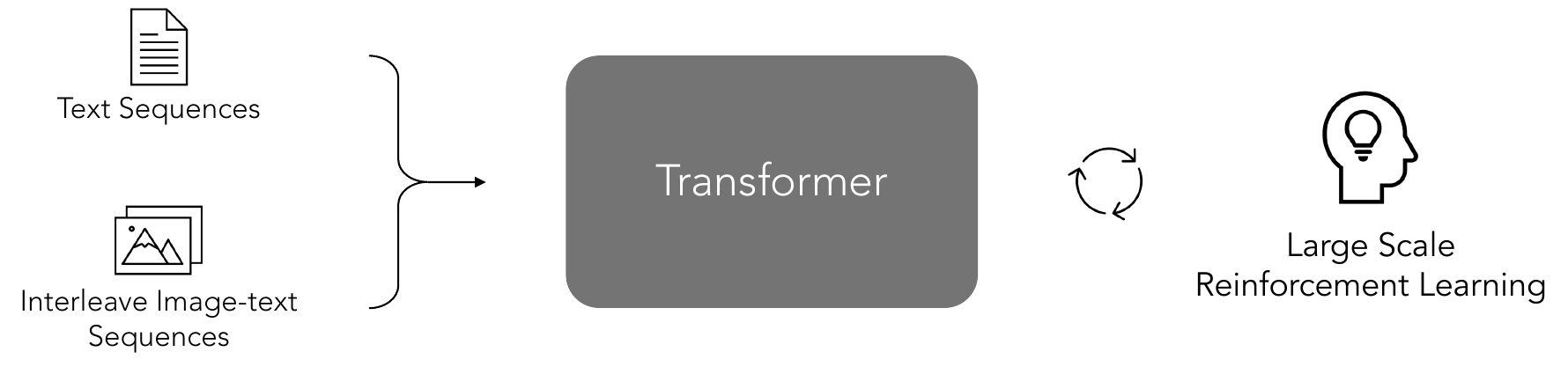}
\caption{Kimi k1.5 supports interleaved images and text as input, leveraging large-scale reinforcement learning to enhance the model's reasoning capabilities.}
\label{fig:arch}
\end{figure}

Kimi k-series models employ a variant of the Transformer decoder \citep{transformer} that integrates multimodal capabilities alongside improvements in architecture and optimization strategies, illustrated in Figure \ref{fig:arch}. These advancements collectively support stable large-scale training and efficient inference, tailored specifically to large-scale reinforcement learning and the operational requirements of Kimi users.

Extensive scaling experiments indicate that most of the base model performance comes from improvements in the quality and diversity of the pretraining data. Specific details regarding model architecture scaling experiments lie beyond the scope of this report and will be addressed in future publications.

\subsection{Training Stages}
The Kimi k1.5 model is trained in three stages: the vision-language pretraining stage, the vision-language cooldown stage, and the long-context activation stage. Each stage of the Kimi k1.5 model’s training focuses on a particular capability enhancement. 

\paragraph{Vision-language pretraining stage} In this stage, the model is firstly trained solely on language data, establishing a robust language model foundation. 
Then the model is gradually introduced to interleaved vision-language data, acquiring multimodal capabilities. The visual tower is initially trained in isolation without updating the language model parameters, then we unfreeze the language model layers, and ultimately increase the proportion of vision-text data to 30\%. The final data mixtures and their respective weights were determined through ablation studies conducted on smaller models. 

\paragraph{Vision-language cooldown stage}
The second stage serves as a cooldown phase, where the model is continue trained with high-quality language and vision-language datasets to ensure superior performance. 
Through empirical investigation, we observed that the incorporation of synthetic data during the cooldown phase yields significant performance improvements, particularly in mathematical reasoning, knowledge-based tasks, and code generation.
The English and Chinese components of the cooldown dataset are curated from high-fidelity subsets of the pre-training corpus. For math, knowledge, and code domains, we employ a hybrid approach: utilizing selected pre-training subsets while augmenting them with synthetically generated content. Specifically, we leverage existing mathematical, knowledge and code corpora as source material to generate question-answer pairs through a proprietary language model, implementing rejection sampling techniques to maintain quality standards~\citep{yue2023mammoth,su2024nemotron}. These synthesized QA pairs undergo comprehensive validation before being integrated into the cooldown dataset.

\paragraph{Long-context activation stage}
Finally, in the third stage, k1.5 is trained with upsampled long-context cooldown data, enabling it to process extended sequences and support tasks that demand longer context. To ensure excellent long-text capabilities of the base model, we upsampled long-context data and used 40\% full attention data and 60\% partial attention data during long context training. The full attention data came partly from high-quality natural data and partly from synthetic long context Q\&A and summary data. The partial attention data came from uniform sampling of cooldown data.
The RoPE frequency~\citep{su2024roformer} was set to 1,000,000. During this stage, we gradually extended length activation training by increasing the maximum sequence length from 4,096 to 32,768, and ultimately to 131,072.

\section{Evaluation Details}
\label{sec:appendix_eval_detail}

\subsection{Text Benchmark}

\textbf{MMLU}~\citep{Hendrycks2020MeasuringMM} covers 57 subjects in STEM, the humanities, social sciences, and more. It ranges in difficulty from an elementary level to an advanced professional level, and it tests both world knowledge and problem-solving ability.

\textbf{IF-Eval}~\citep{Zhou2023InstructionFollowingEF} is a benchmark for evaluating large language models' ability to follow verifiable instructions. There are 500+ prompts with instructions such as "write an article with more than 800 words", etc.
Due to a version shift, the number of IFEval reported in Table \ref{tab:short_perf} derived from an intermediate model. We will update the scores based on the final model.

\textbf{CLUEWSC}~\citep{Xu2020CLUEAC} is a coreference resolution task in CLUE benchmark, requiring models to determine if a pronoun and a noun phrase in a sentence co-refer, with data from Chinese fiction books.

\textbf{C-EVAL}~\citep{Huang2023CEvalAM} is a comprehensive Chinese evaluation suite for assessing advanced knowledge and reasoning abilities of foundation models. It includes 13,948 multiple-choice questions across 52 disciplines and four difficulty levels.

\subsection{Reasoning Benchmark}

\textbf{HumanEval-Mul} is a subset of Multipl-E~\citep{Cassano2022MultiPLEAS}. MultiPL-E extends the HumanEval benchmark and MBPP benchmark to 18 languages that encompass a range of programming paradigms and popularity. We choose HumanEval translations in 8 mainstream programming languages (Python, Java, Cpp, C\#, JavaScript, TypeScript, PHP, and Bash).

\textbf{LiveCodeBench}~\citep{Jain2024LiveCodeBenchHA} serves as a comprehensive and contamination-free benchmark for assessing large language models (LLMs) in coding tasks.  It features live updates to prevent data contamination, holistic evaluation across multiple coding scenarios, high-quality problems and tests, and balanced problem difficulty. We test short-CoT model with questions from 2408-2411 (release v4), and long-CoT model with questions from 2412-2502 (release v5).

\textbf{AIME 2024} comprises the competition questions for the AIME in 2024. The AIME is a prestigious, invitation-only math contest for top high school students, assessing advanced math skills and requiring solid foundation and high logical thinking.

\textbf{MATH-500}~\citep{lightman2023lets} is a comprehensive mathematics benchmark that contains 500 problems on various mathematics topics including algebra, calculus, probability, and more. Tests both computational ability and mathematical reasoning. Higher scores indicate stronger mathematical problem-solving capabilities.

\textbf{Codeforces} is a well-known online judge platform and serves as a popular testbed for evaluating long-CoT coding models. To achieve higher rankings in the Div2 and Div3 competitions, we utilize majority voting on the code snippets generated by the k1.5 long-CoT model, employing test cases that are also generated by the same model. The percentile of the codeforce ELO rating was extracted from OpenAI Day12 talk \footnote{\url{https://www.youtube.com/watch?v=SKBG1sqdyIU&ab_channel=OpenAI}}c.

\subsection{Image Benchmark}

\textbf{MMMU}~\citep{yue2024mmmu} encompasses a carefully curated collection of 11.5K multimodal questions sourced from college exams, quizzes, and textbooks. These questions span six major academic fields: Art \& Design, Business, Science, Health \& Medicine, Humanities \& Social Science, and Tech \& Engineering.

\textbf{MATH-Vision} (MATH-V)~\citep{wang2024measuring} is a carefully curated collection of 3,040 high-quality mathematical problems with visual contexts that are sourced from real math competitions. It covers 16 distinct mathematical disciplines and is graded across 5 levels of difficulty. This dataset offers a comprehensive and diverse set of challenges, making it ideal for evaluating the mathematical reasoning abilities of LMMs.

\textbf{MathVista}~\citep{lu2023mathvista} is a benchmark that integrates challenges from a variety of mathematical and visual tasks, demanding participants to exhibit fine-grained, deep visual understanding along with compositional reasoning to successfully complete the tasks.

\end{document}